\documentclass[sigconf]{acmart}

\copyrightyear{2026}
\acmYear{2026}
\setcopyright{cc}
\setcctype{by}
\acmConference[WWW '26]{Proceedings of the ACM Web Conference 2026}{April 13--17, 2026}{Dubai, United Arab Emirates}
\acmBooktitle{Proceedings of the ACM Web Conference 2026 (WWW '26), April 13--17, 2026, Dubai, United Arab Emirates}
\acmPrice{}
\acmDOI{10.1145/3774904.3792357}
\acmISBN{979-8-4007-2307-0/2026/04}
\usepackage{hyperref}       % hyperlinks
\usepackage{url}            % simple URL typesetting
\usepackage{booktabs}       % professional-quality tables
\usepackage{amsfonts}       % blackboard math symbols
\usepackage{makecell}
\usepackage{nicefrac}       % compact symbols for 1/2, etc.
\usepackage{microtype}      % microtypography
\usepackage{xcolor}         % colors
\usepackage{xpatch}
\usepackage{wrapfig}
\usepackage{amsmath}
\usepackage{algorithm}
\usepackage[noend]{algorithmic}
\usepackage{amsmath}
\usepackage{multirow}
\usepackage{xpatch}
\usepackage{graphicx}
\usepackage[most]{tcolorbox}
\usepackage{color}
\usepackage{colortbl}
\usepackage{subcaption} % 子图

\begin{document}

\title{EMSEdit: Efficient Multi-Step Meta-Learning-based Model Editing}


\author{Xiaopeng Li}
\orcid{0009-0008-8695-5591}
\affiliation{\institution{National University of Denfense Technology}
\city{Changsha}
\country{China}}
\email{xiaopengli@nudt.edu.cn}
\author{Shasha Li{\fontsize{9.5pt}{\baselineskip}\selectfont *}}
\orcid{0000-0002-6508-5119}
\affiliation{\institution{National University of Denfense Technology}
\city{Changsha}
\country{China}}
\email{shashali@nudt.edu.cn}
\author{Xi Wang}
\orcid{0009-0005-7668-3965}
\affiliation{\institution{National University of Denfense Technology}
\city{Changsha}
\country{China}}
\email{wx2023@nudt.edu.cn}
\author{Shezheng Song}
\orcid{0009-0007-9985-7619}
\affiliation{\institution{National University of Denfense Technology}
\city{Changsha}
\country{China}}
\email{ssz614@nudt.edu.cn}
\author{Bin Ji}
\orcid{0000-0002-5508-5051}
\affiliation{\institution{National University of Denfense Technology}
\city{Changsha}
\country{China}}
\email{jibin@nudt.edu.cn}
\author{Shangwen Wang}
\orcid{0000-0003-1469-2063}
\affiliation{\institution{National University of Denfense Technology}
\city{Changsha}
\country{China}}
\email{wangshangwen13@nudt.edu.cn}
\author{Jun Ma}
\orcid{0000-0003-2258-0854}
\authornote{Shasha Li, Jun Ma, and Jie Yu are the corresponding authors.}
\affiliation{\institution{National University of Denfense Technology}
\city{Changsha}
\country{China}}
\email{majun@nudt.edu.cn}
\author{Xiaodong Liu}
\orcid{0000-0002-9800-6886}
\affiliation{\institution{National University of Denfense Technology}
\city{Changsha}
\country{China}}
\email{liuxiaodong@nudt.edu.cn}
\author{Mina Liu}
\orcid{0009-0006-9554-0772}
\affiliation{\institution{KylinSoft}
\city{Changsha}
\country{China}}
\email{a506lm@126.com}
\author{Jie Yu{\fontsize{9.5pt}{\baselineskip}\selectfont *}}
\orcid{0009-0007-1545-7010}
\affiliation{\institution{National University of Denfense Technology}
\city{Changsha}
\country{China}}
\email{yj@nudt.edu.cn}

\renewcommand{\shortauthors}{Xiaopeng Li  et al.}

\begin{abstract}
Large Language Models (LLMs) power numerous AI applications, yet updating their knowledge remains costly. Model editing provides a lightweight alternative through targeted parameter modifications, with meta-learning-based model editing (MLME) demonstrating strong effectiveness and efficiency. However, we find that MLME struggles in low-data regimes and incurs high training costs due to the use of KL divergence. To address these issues, we propose \textbf{E}fficient \textbf{M}ulti-\textbf{S}tep \textbf{Edit (EMSEdit)}, which leverages multi-step backpropagation (MSBP) to effectively capture gradient-activation mapping patterns within editing samples, performs multi-step edits per sample to enhance editing performance under limited data, and introduces norm-based regularization to preserve unedited knowledge while improving training efficiency. Experiments on two datasets and three LLMs show that EMSEdit consistently outperforms state-of-the-art methods in both sequential and batch editing. Moreover, MSBP can be seamlessly integrated into existing approaches to yield additional performance gains. Further experiments on a multi-hop reasoning editing task demonstrate EMSEdit's robustness in handling complex edits, while ablation studies validate the contribution of each design component. Our code is available at \url{https://github.com/xpq-tech/emsedit}.
\end{abstract}



\maketitle
\section{Introduction}
Large Language Models (LLMs) have emerged as foundational infrastructure for a wide range of AI applications \cite{zhao2023survey,liu2025advances}. Through extensive pretraining on diverse corpora followed by alignment, LLMs acquire rich world knowledge. However, the knowledge encoded in LLMs becomes fixed after training, making them unable to adapt to updates in the real world \cite{yao-etal-2023-editing}. Re-training LLMs to reflect new knowledge is prohibitively expensive \cite{biderman2023pythia}. This issue is particularly critical in the context of the Web, which serves as the primary medium for real-time knowledge dissemination. While the Web evolves on a daily basis, LLMs remain static, leading to a growing gap between model knowledge and current information.

\begin{figure*}[t]
    \centering
    \includegraphics[width=0.99\linewidth]{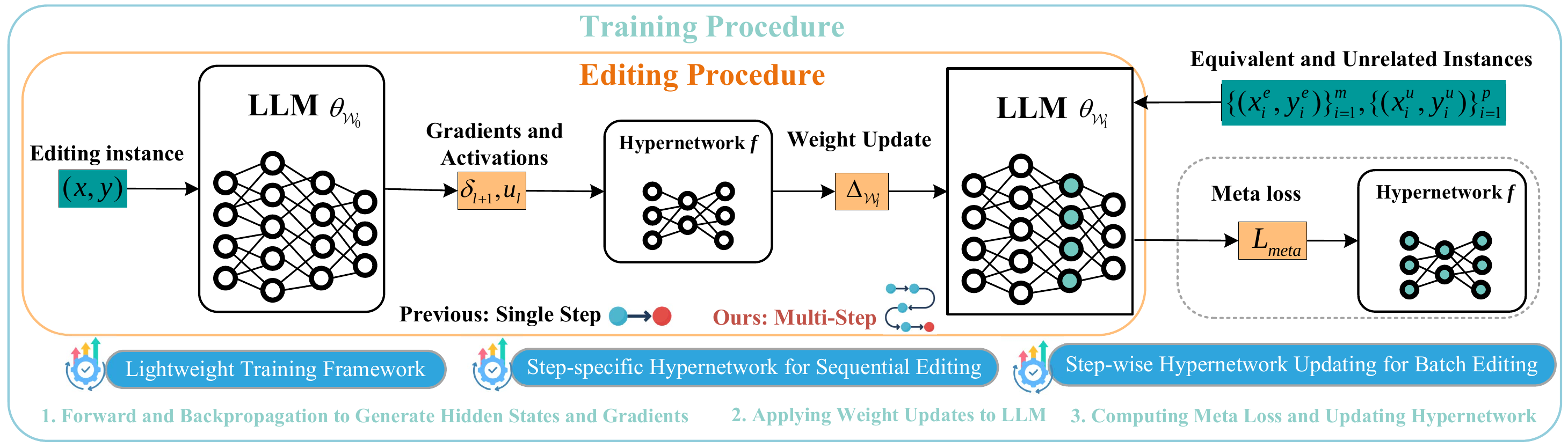}
    \caption{Overview of EMSEdit. Existing MLME methods perform model editing in a single step (i.e., one-step backpropagation), whereas EMSEdit conducts editing through multi-step updates, similar to multi-step backpropagation. %EMSEdit introduces a Lightweight Training Framework, a Step-specific Hypernetwork for Sequential Editing, and a Step-wise Hypernetwork Updating mechanism for Batch Editing to balance editing efficiency and performance.
    }
    \label{fig:ms_method}
\end{figure*}
To address this challenge, recent studies have introduced \textbf{model editing} as an efficient paradigm for updating knowledge in LLMs \cite{zhu2020modifying,Meng2022Locating,mitchell2022fast,fang2024alphaedit,li2025reinforced,wang2024editing}. These approaches aim to make LLMs memorize new information by modifying only a small subset of parameters. Unlike vanilla fine-tuning, model editing typically computes a weight update that indirectly injects new knowledge into the model. For example, \textbf{locate-then-edit} methods perform least-squares updates on the feed-forward networks that are believed to store factual knowledge \cite{meng2022massediting,li2025rethinking,geva-etal-2021-transformer,geva-etal-2023-dissecting}, while \textbf{meta-learning-based model editing (MLME)} employs a lightweight hypernetwork to transform fine-tuning gradients and internal representations into effective weight updates \cite{tan2024massiveeditinglargelanguage,li2025reinforced,mitchell2022fast}. Although locate-then-edit approaches achieve impressive editing performance, they suffer from low efficiency. For instance, in sequential editing tasks, AlphaEdit \cite{fang2024alphaedit} (a representative locate-then-edit method) is approximately 27.8 $\times$ slower than RLEdit \cite{li2025reinforced}, a representative MLME method. MLME approaches exhibit promising performance both in terms of effectiveness and efficiency.

Despite the progress made in MLME, we observe that it still has a couple of limitations. First, current MLME methods perform well when abundant training data is available, but their performance is suboptimal under low-data scenarios (Section \ref{sec:Data Efficiency Limitation}). This highlights a data efficiency limitation in existing MLME methods. Second, the training efficiency of MLME is bottlenecked by the computation of KL divergence loss (Section \ref{sec:Training Time Analysis of MLME}), which requires the reference probability distribution from the original model, leading to double forward passes per iteration.

To tackle the above issues, we propose \textbf{Efficient Multi-Step Edit (EMSEdit)}, which performs model editing through multi-step
edits on each editing sample, leveraging multi-step backpropagation (MSBP) to sufficiently learn editing patterns from the limited training data. This enables more effective editing with limited data, thereby addressing the first issue. To improve training efficiency, EMSEdit replaces the KL loss with an $l_2$ regularization on the weight update to preserve the original model’s knowledge, a design we term the lightweight training framework, thus alleviating the second issue. Furthermore, EMSEdit introduces a step-specific hypernetwork for sequential editing and a step-wise hypernetwork updating mechanism for batch editing, balancing editing effectiveness and efficiency. The comparison between EMSEdit and existing MLME methods is illustrated in Figure~\ref{fig:ms_method}. 

We evaluate EMSEdit on both batch and sequential editing tasks using GPT-J (6B) \cite{gpt-j}, LLaMA-3 (8B) \cite{llama3modelcard}, and Gemma-2 (9B) \cite{team2024gemma} on the ZsRE \cite{zhu2020modifying} and CounterFact \cite{Meng2022Locating} datasets. Experimental results demonstrate that \textbf{EMSEdit surpasses state-of-the-art methods in both sequential and batch editing}. We further validate that EMSEdit achieves superior performance on the multi-hop reasoning editing task, and that \textbf{incorporating MSBP consistently enhances the editing performance of existing MLME baselines}. In addition, we validate the effectiveness of the components introduced in EMSEdit through ablation studies and further demonstrate its data and memory efficiency through additional analyses. In summary, our contributions are as follows:
\begin{itemize}
    \item We reveal that current MLME methods face limitations in both data and training efficiency. Their performance is suboptimal under low-data scenarios and training efficiency is bottlenecked by the computation of KL loss.
    \item We propose to use MSBP and norm constraints to address the limitations of existing MLME methods. We thus introduce EMSEdit, a novel MLME approach that effectively balances editing performance and efficiency through the integration of MSBP and three key designs.
    \item We demonstrate the effectiveness of EMSEdit and MSBP through sequential and batch editing experiments on three LLMs and two datasets. Our additional experiments and analyses respectively demonstrate that EMSEdit remains effective on complex multi-hop reasoning editing tasks and exhibits superior data and memory efficiency.

\end{itemize}
\section{Related Work}
Model editing has recently emerged as a promising technique for updating the knowledge of large language models (LLMs) \cite{zhang2024comprehensive}. These approaches can be broadly categorized into two types based on whether they modify the parameters of LLMs: \textbf{parameter-altering} and \textbf{parameter-preserving} methods \cite{yao-etal-2023-editing}.

\textbf{Parameter-preserving} methods store editing information externally—either by retrieving relevant knowledge from external sources \cite{zheng-etal-2023-edit}, introducing auxiliary modules outside the model \cite{mitchell2022memory}, or adding new internal parameters \cite{wang2024wiserethinkingknowledgememory,hartvigsen2024aging}. These approaches have shown promising results in model editing tasks while mitigating the risks of altering the model's original behavior. However, the external modules often increase system complexity and computational overhead, limiting their practicality in real-world applications.

In contrast, \textbf{parameter-altering} methods produce edited models that differ from the original ones only in their parameters \cite{zhu2020modifying,Meng2022Locating,mitchell2022fast}. This design makes them inherently more lightweight and flexible, allowing the edited models to be further edited or adapted to downstream tasks seamlessly. Such methods typically perform editing by modifying a small subset of the LLM’s parameters. For example, locate-then-edit methods apply least-squares updates to key layers believed to store factual knowledge \cite{meng2022massediting,fang2024alphaedit,li2025rethinking}, achieving promising editing results. However, these methods often suffer from limited editing efficiency.

\textbf{Meta-learning-based model editing (MLME)} represents a class of approaches that are competitive in both effectiveness and efficiency \cite{tan2024massiveeditinglargelanguage,li2025reinforced}. These methods train a lightweight hypernetwork to transform single-step fine-tuning gradients and internal representations into effective weight updates \cite{mitchell2022fast}. In this paper, we propose a novel MLME method, EMSEdit. Unlike existing MLME methods, EMSEdit leverages MSBP to make fuller use of the editing data, thereby enabling more expressive and effective model editing.
\section{Preliminaries}
\subsection{Model Editing Problem}\label{sec:me_problem}
Model editing aims to efficiently modify a LLM $\theta$ to achieve specific editing goals without affecting its other capabilities. We define an editing sample as
\begin{equation}
 T = (x, y) \cup \{(x^e_i, y^e_i)\}_{i=1}^m \cup \{(x^u_i, y^u_i)\}_{i=1}^p
\end{equation}
where $(x, y)$ is the editing instance used to measure the \textbf{efficacy} of model editing methods; ${(x^e_i, y^e_i)}{i=1}^m$ are examples that are semantically equivalent to $(x, y)$ and are used to evaluate the \textbf{generalization} ability; and ${(x^u_i, y^u_i)}{i=1}^p$ are semantically unrelated examples used to assess the \textbf{specificity} of the model editing methods.
Additionally, to assess whether the model retains its original capabilities after editing, existing works evaluate the performance of edited models on downstream natural language processing tasks (e.g., GLUE \cite{wang2018glue}). Model editing methods can be classified into three scenarios: \textbf{sequential editing}, \textbf{batch editing}, and \textbf{sequential batch editing} \cite{mazzia2023survey}.
Given $n$ editing samples $\{T_1, T_2, \dots, T_n\}$ to be edited, the three scenarios can be described as follows: \textbf{a) Sequential Editing:} This scenario continuously edits a single piece of knowledge, processing the $n$ editing samples one by one; \textbf{ b) Batch Editing:} This scenario performs a single edit operation on all $n$ samples simultaneously, effectively editing multiple pieces of knowledge at once; \textbf{c) Sequential Batch Editing:} This scenario divides the $n$ samples into batches of size $b$ and performs $\left\lceil \frac{n}{b} \right\rceil$ sequential editing steps, each handling a batch.

After completing the editing process, the performance of the editing method is evaluated by measuring the success rate of the edited model across the $n$ editing samples. The ultimate goal of all three scenarios is to effectively write the $n$ knowledge samples into $\theta$. However, their applicable scenarios, approaches to achieving the goal, and implementation challenges differ significantly.
\subsection{Meta Learning Based Model Editing}\label{sec:Meta Learning Based Model Editing}
Meta-learning-based model editing trains a hypernetwork to convert standard fine-tuning gradients into weight updates, reducing the computational burden of direct weight adjustment \cite{mitchell2022fast,tan2024massiveeditinglargelanguage,li2025reinforced}.

\textbf{Editing Procedure:} We denote the weight of $\theta$ as: $\mathcal{W} = \{W_l: l\in \mathcal{L}\}$ where $W_l$ is the weight of linear layer $l$ and $\mathcal{L}$ is the set of all linear layers. MEND \cite{mitchell2022fast} decomposes the single BP gradient $\nabla_{W_l}$ of the weight $W_l\in \mathbb{R}^{d'\times d}$ in layer $l$ to be updated into a rank-1 product: $ \nabla_{W_l}=\delta_{l+1}u_{l}^{T}$, where $\delta_{l+1} \in \mathbb{R}^{d'\times 1}$ is the gradient of loss $L$ with respect to the preactivations to layer $l+1$ and $u_{l}\in \mathbb{R}^{d\times 1}$ are the inputs to layer $l$. The weight update $\Delta_{W_l}\in \mathbb{R}^{d'\times d}$ is obtained as the product of the \textit{pseudogradient} and \textit{pseudoactivations} transformed by a hypernetwork:
\begin{equation}
    \Delta_{W_l} = \tilde\delta_{l+1}\tilde u_{l}^{T}, \text{where  } \tilde \delta_{l+1},\tilde u_{l}^{T} =
    f(\delta_{l+1},u_{l}^{T})
\end{equation} where $f: \mathbb{R}^d \times \mathbb{R}^{d'} \rightarrow \mathbb{R}^d \times \mathbb{R}^{d'}$ is the hypernetwork which transforms the gradient and inputs to \textit{pseudogradient} $\tilde \delta_{l+1}$ and \textit{pseudoactivations} $\tilde u_{l}^{T}$. MALMEN \cite{tan2024massiveeditinglargelanguage} improves the above process to accommodate massive batch editing scenarios by formulating a least squares problem that aggregates the weight updates corresponding to each editing sample into a single weight update. The weight update $\Delta_{W_l}$ of MALMEN is:
\begin{equation}
\begin{aligned}
    \Delta_{W_l} &= D_l U_l^T(U_lU_l^T + \lambda_l I)^{-1} \\&\text{  by solving: } \mathop{\min}_{\Delta_{W_l} \in \mathbb{R}^{d' \times d}}|| \Delta_{W_l} U_l - D_l||^2_2 + \lambda ||\Delta_{W_l}||^2_2
\end{aligned}
\end{equation} where $D_l = [...,d_{l,k},...] = [..., \tilde \delta_{l+1,k}\tilde u_{l,k}^{T}u_{l,k},...] \in \mathbb{R}^{d'\times b}$ and $U_l = [...,u_{l,k},...]\in \mathbb{R}^{d\times b}$, with $k$ representing the $k$-th editing sample in a batch of size $b$.

\textbf{Training Procedure:} The purpose of this process is to obtain a hypernetwork $f$ that can transform gradients and activations into weight updates capable of achieving the model editing objective. The goal of model editing is to perform the edit while maintaining good generalization and specificity, which corresponds to the following two loss functions:
\begin{equation}
        L_e = - \log \theta_{\mathcal{W}_1} (y^e|x^e), L_{loc} = \text{KL}\left[\theta_{\mathcal{W}_0} (\cdot|x^u)||\theta_{\mathcal{W}_1} (\cdot|x^u) \right]
\end{equation}where $\theta_{\mathcal{W}_0}$ and $\theta_{\mathcal{W}_1}$ denote the LLM with the original weights and the updated weights after one edit, respectively. The meta loss for training $f$ is the sum of $L_e$ and $L_{loc}$: $L_{meta} = L_e + \lambda_{loc} L_{loc}$ where $\lambda_{loc}$ is a hyperparameter that balances generalization and specificity.

Based on the meta loss, different methods adopt various implementation details. MEND back-propagates $L_{meta}$ into $f$ via LLM $\theta$ \cite{mitchell2022fast}, while MALMEN back-propagates on $\theta$ and $f$ separately to avoid high memory costs \cite{tan2024massiveeditinglargelanguage}. RLEdit introduces a reinforcement learning framework into the above process, thus enabling the hypernetwork to perform sequential model editing in $\theta$ \cite{li2025reinforced}. Specifically, RLEdit optimizes the following reward to obtain the optimal parameters of $f$:
\begin{equation}
    J = \mathop{\sum}_{i=1}^n \gamma^i(L_{meta_i} + L_{back_i} + \eta||\Delta_{\mathcal{W}_i}||^2)
\end{equation} where the term $\eta||\Delta_{\mathcal{W}_i}||^2$ is the regularization term, with coefficient $\eta$, that accounts for the weight updates $\Delta_{\mathcal{W}_i}$ at edit step $i$. $\gamma^i$ is the discount factor. The term $L_{back_i}$ is the memory backtracking loss. RLEdit computes the meta loss of the current  editing sample $T_i$ and the meta loss of previous $q$ editing samples $\{T_{i-1},...,T_{i-q}\}$ on $\theta_{\mathcal{W}_{i-1}}$ at edit step $i$. The backtracking loss is defined as:
    \begin{equation}
        L_{back_i} = \mathop{\sum}_{j=i-q}^{i-1}\mu ^{i-j}(L_{e_j,\mathcal{W}_{i-1}} + \lambda_{loc}L_{loc_j,\mathcal{W}_{i-1}})
    \end{equation}where $\mu$ is a decay factor.

\section{Limitations of MLME}\label{sec:analyze}
In this study, we use two state-of-the-art MLME methods as research subjects and baselines: RLEdit
\cite{li2025reinforced} for sequential editing and MALMEN \cite{tan2024massiveeditinglargelanguage} for batch editing.
\subsection{Data Efficiency Limitation}\label{sec:Data Efficiency Limitation}
\begin{figure}[t]
    \centering
    \includegraphics[width=0.85\linewidth]{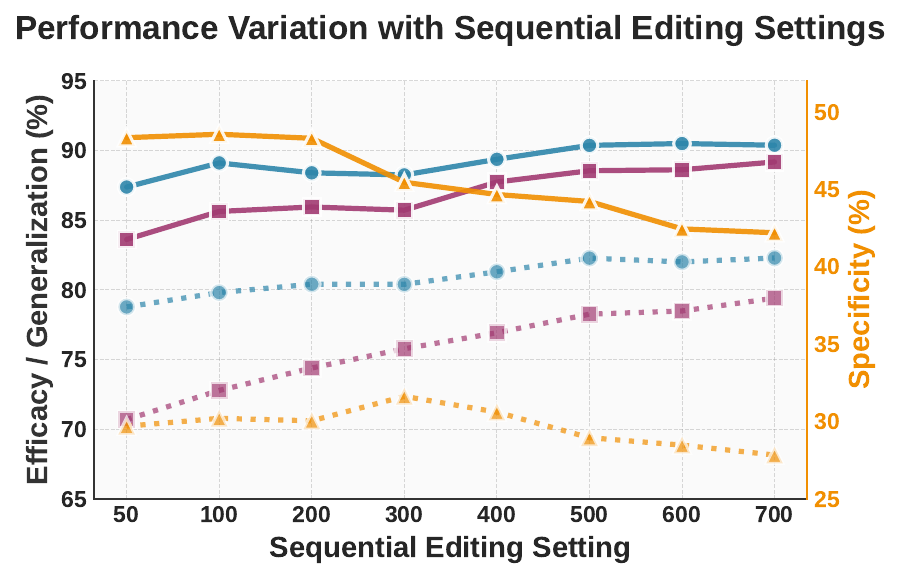}
    \includegraphics[width=0.9\linewidth]{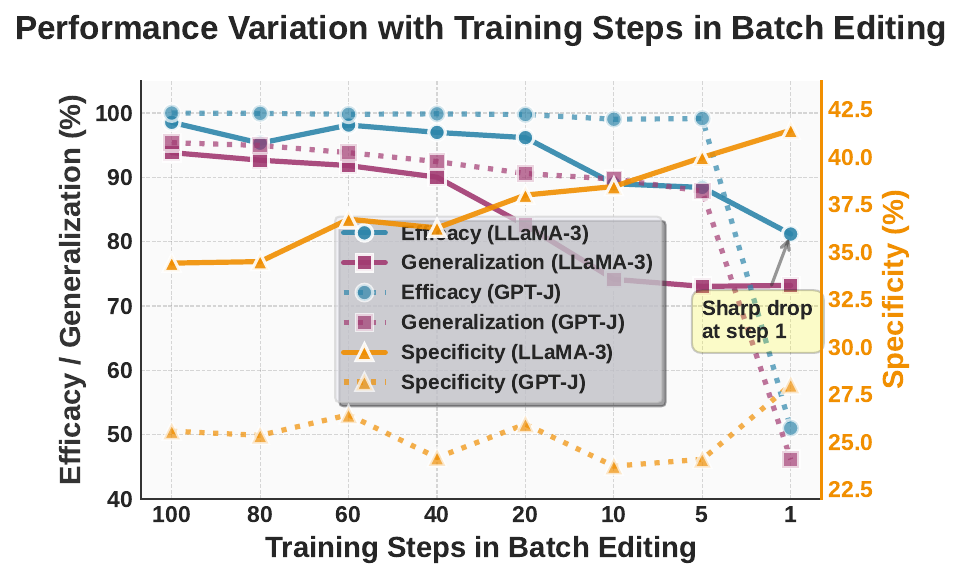}
    \caption{Performance Variation of MLME in Sequential and Batch Editing. The batch size for sequential editing is 20.}
    \label{fig:es_compare}
\end{figure}
We observe that MLME’s performance follows a saturating growth trend with respect to the amount of training data it receives. In Figure \ref{fig:es_compare}, we show how MLME’s editing performance varies with the number of sequential edits and training steps in batch edits.

In the sequential editing setting\footnote{In our sequential editing experiments with RLEdit, the number of training samples is matched to the number of sequential edits, following the same setup as in \cite{li2025reinforced}.}, the overall editing performance increases as the number of sequential edits grows. This result is somewhat counterintuitive, as one might expect sequential model editing to become more challenging with a greater number of edits. We hypothesize that longer editing sequences expose the model to more training instances, thereby enhancing its performance.

To test this hypothesis, we conduct batch editing experiments involving 1,000 batch edits, varying the number of training steps to simulate different amounts of training data. At each step, the number of training samples is equal to the batch size, and the training data differed across steps. As shown in Figure~\ref{fig:es_compare} (bottom), editing performance decreases as the number of training steps is reduced. Especially, when the number of training steps is 1—meaning the amount of training data is equal to the batch size—the editing performance of MLME drops significantly. This further supports our hypothesis, leading to the conclusion that \textbf{the performance of MLME improves with increased training data, and that when training data is scarce, MLME suffers a substantial performance degradation.} Given that training data is often scarce or hard to acquire in real-world applications \cite{10.1093/nsr/nwx110}, the \textbf{current design of MLME yields suboptimal editing performance in low-resource settings}, thereby constraining its practical utility.
\subsubsection{Solutions to Mitigate the Data Efficiency Limitation}\label{sec: Solutions to Mitigate the Data Efficiency Limitation}
The most direct way to mitigate the limitations of MLME is to increase the amount of training data. However, this approach is difficult to implement in data-scarce scenarios. Another option is to modify the training strategy of MLME. We observe that the current MLME performs model editing using only a single-step gradient, which is equivalent to performing single backpropagation. \begin{figure}[t]
    \centering
    \includegraphics[width=0.99\linewidth]{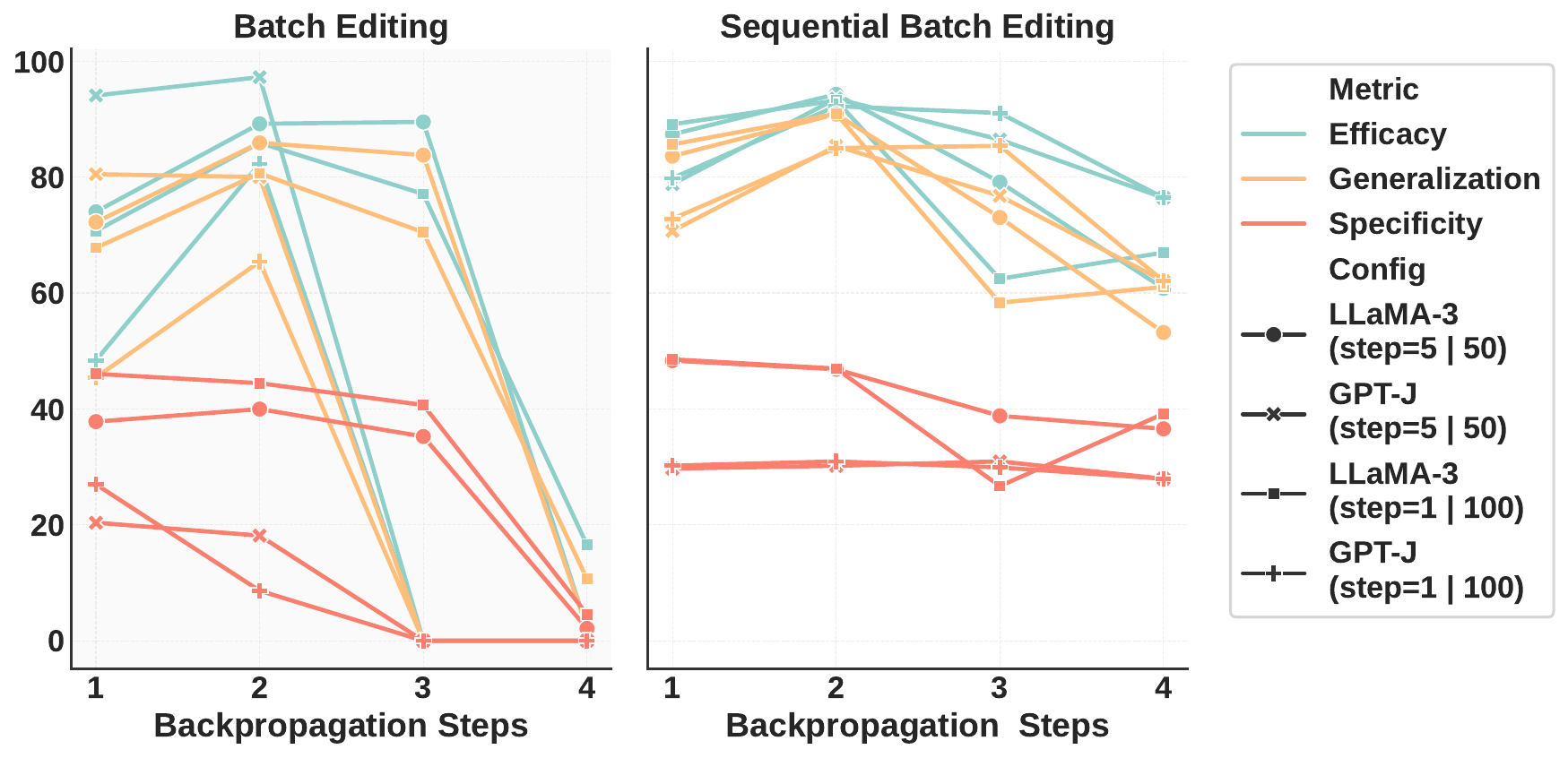}
    \caption{Performance Variation of MLME in Sequential and Batch Editing Across Different BP Steps on ZsRE Dataset. (step=5 | 50) means the training steps is 5 in batch editing and sequence number is 50 in sequential editing.}
    \label{fig:iteration}
\end{figure}As a result, MLME may not fully capture the editing patterns in the training process. To verify this, we modify MLME to adopt a multi-step gradient-based approach, where multi-step backpropagation (MSBP) is performed on a batch of data to achieve model editing.

We show the performance variation of MLME in both batch and sequential batch editing settings across different numbers of BP steps in Figure \ref{fig:iteration}. In the batch editing setting (left), we set the number of batches as 8,192. We limit the number of training steps to simulate the scarcity of training data in real-world scenarios. The efficacy and generalization of MLME generally exhibit a trend of first increasing and then decreasing, while specificity consistently decreases as the number of backpropagation (BP) steps increases. In the sequential batch editing (right figure in Figure \ref{fig:iteration}, batch=30),  the trends observed here are similar to those in batch editing, but the best editing performance is achieved when the step size is 2. This result is intuitive: more BP steps mean more updates, and performing multiple updates on a single batch can lead to overfitting to that batch, thereby increasing efficacy and generalization but reducing specificity. Moreover, when the number of BP steps exceeds a certain threshold, the editing performance drops sharply, which may be due to excessive updates causing the editing collapse. This suggests that \textbf{MSBP can indeed achieve better editing performance}, and the choice of step size is a critical factor, with a step size of 2 yielding the overall best results.
\subsection{Training Efficiency Limitation of MLME}\label{sec:Training Time Analysis of MLME}
Although MLME demonstrates strong editing efficiency, its training efficiency remains suboptimal. To investigate this issue, we conduct an in-depth analysis of the training time distribution for two representative MLME methods, MALMEN and RLEdit, under both batch editing and sequential batch editing settings. Specifically, we perform batch editing with a batch size of 8,192 and sequential batch editing with 30$\times$200 updates using MALMEN and RLEdit on the ZsRE and CounterFact datasets, employing GPT-J and LLaMA-3 (8B) as the underlying models. We measure the average time consumption of the five steps in the MLME training process:
1. Caching $\nabla_{W_l}$;
2. Computing $\Delta_{W_l}$;
3. Computing $L_e$ and BP;
4. Computing $L_{loc}$ and BP;
5. Updating $f$.

\begin{figure}[t]
    \centering
    \includegraphics[trim=6 15 6 37,clip, width=0.9\linewidth]{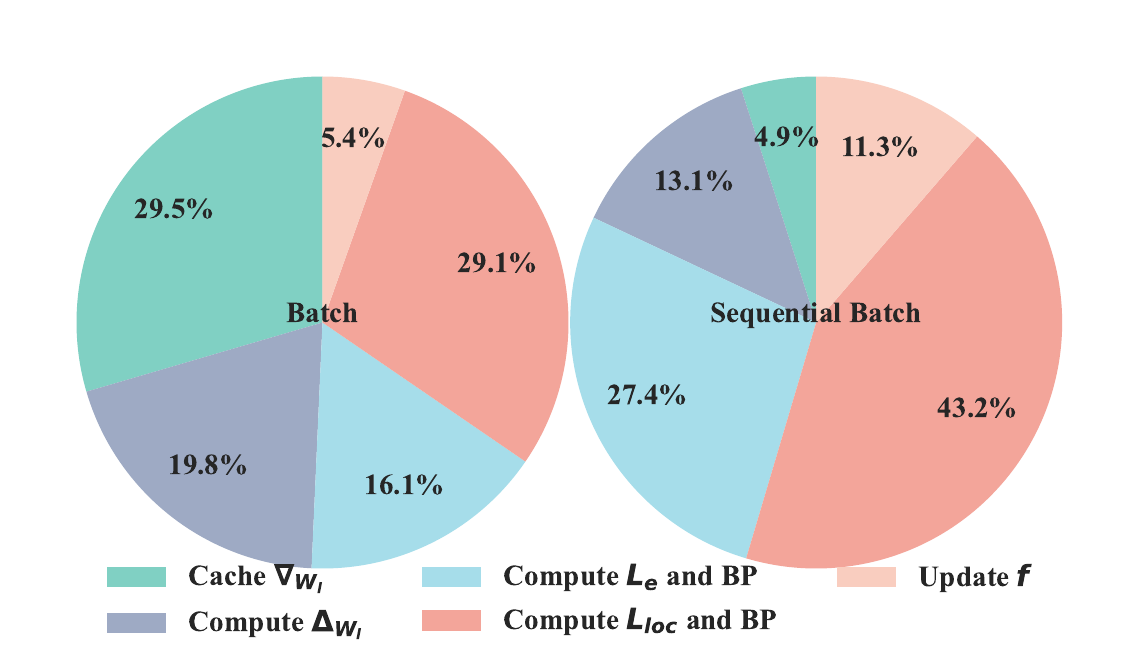}
    \caption{\small MLME's Time Distribution.}
    \label{fig:time_consume}
\end{figure}
Figure~\ref{fig:time_consume} shows the time distribution of MLME. In both the Batch and Sequential Batch experiments, Step 4 (computing $L_{loc}$ and BP) accounts for a significant portion of the total training time—29.1\% and 43.2\%, respectively. This step is designed to protect the model's original knowledge. This overhead is expected because calculating the KL loss requires two forward passes through the LLMs (one for the original model output and one for the edited model output), leading to increased computational cost. Although Step 1 (Cache $\nabla_{W_l}$) consumes more time than Step 4 in the Batch Editing setting, it involves collecting the original fine-tuning gradients and internal representations, which is a fundamental step in MLME. Notably, its time cost is the lowest in Sequential Batch Editing, indicating that it is not the primary bottleneck.

From the above analysis, we can conclude that \textbf{the training bottleneck of MLME lies in Step 4: Compute $\boldsymbol{L_{loc}}$ and BP.} Therefore, we need to find an alternative way to efficiently preserve the model's original knowledge. To achieve this, we introduce the $l_2$-norm constraint as an alternative in Section \ref{sec:Lightweight Training}, and demonstrate in the batch editing and sequential editing experiments (Section \ref{sec:exp}) that this alternative can achieve the aforementioned goal.

\section{Method}
Following the same editing pipeline as MLME, \textbf{E}fficient \textbf{M}ulti-\textbf{S}tep \textbf{Edit} (\textbf{EMSEdit}) consists of an \textbf{\textit{Editing Procedure}} and a \textbf{\textit{Training Procedure}}. However, EMSEdit incorporates \textit{MSBP-based editing, step-specific hypernetwork for sequential editing, step-wise hypernetwork updating for batch editing, and an efficient and lightweight training process}. We describe them in detail below.
\begin{algorithm}[t]
\small
   \caption{Hypernetwork Training for Sequential Editing}
   \label{alg:smtrain}
   \begin{algorithmic}[1]
   \STATE {\bfseries Input:} LLM $\theta_{\mathcal{W}_0}$, hyperparameters $S$ and $\eta$, set of hypernetworks $\mathcal{F}=\{f_1,...,f_S \}$
   \STATE {\bfseries Output:} Optimized hypernetworks $\mathcal{F}$
   \STATE Sample $(T_i = \{x_i, y_i, x^e_i, y^e_i\})_{i=1}^n$
   \STATE \textcolor{cyan}{// Efficient and lightweight training process}
   \FOR{$t=1$ {\bfseries to} $n$}
    \STATE \textcolor{cyan}{// MSBP}
   \FOR{$s=1$ {\bfseries to} $S$}
   \STATE $L \leftarrow -\log{\theta_{\mathcal{W}^{s-1}_{t-1}}(y_t \mid x_t)}$
   \STATE Back-propagate $L$ and cache $\nabla_{\mathcal{W}_{t-1}}$
   \STATE \textcolor{cyan}{// Step-specific
hypernetwork}
   \STATE $\Delta_{s-1,\mathcal{W}_{t-1}} \leftarrow f_{s-1}(\nabla_{\mathcal{W}_{t-1}})$
   \STATE $\mathcal{W}^s_t \leftarrow \mathcal{W}^{s-1}_{t-1} + \Delta_{s-1,\mathcal{W}_{t-1}}$
   \ENDFOR
   \STATE  $L_{cons} \leftarrow  \|\mathcal{W}^S_t -\mathcal{W}_0 \|^2$
   \STATE $L_{e,i} \leftarrow -\log{\theta_{\mathcal{W}_t}(y^e_i \mid x^e_i)}$
    \STATE $L_i \leftarrow L_{e,i} + \eta L_{cons}$
    \STATE Back-propagate $L_i$
   \ENDFOR
   \STATE Update $\mathcal{F}$
   \STATE \textbf{return} $\mathcal{F}$
   \end{algorithmic}
\end{algorithm}
\begin{algorithm}[t]
\small
   \caption{Hypernetwork Training for Batch Editing}
   \label{alg:smtrain-batch}
   \begin{algorithmic}[1]
   \STATE {\bfseries Input:} LLM $\theta_{\mathcal{W}_0}$, hyperparameters $S$ and $\eta$, a hypernetwork $f$
   \STATE {\bfseries Output:} An optimized hypernetwork $f$
   \STATE Sample $(T_i = \{x_i, y_i, x^e_i, y^e_i\})_{i=1}^n$
   \STATE \textcolor{cyan}{// Efficient and lightweight training process}
   \FOR{$t=1$ {\bfseries to} $n$}
    \STATE \textcolor{cyan}{// MSBP}
   \FOR{$s=1$ {\bfseries to} $S$}
   \STATE \textcolor{cyan}{// Step-wise hypernetwork updating}
   \STATE $L \leftarrow -\log{\theta_{\mathcal{W}^{s-1}_{t-1}}(y_t \mid x_t)}$
   \STATE Back-propagate $L$ and cache $\nabla_{\mathcal{W}_{t-1}}$
   \STATE $\Delta_{s-1,\mathcal{W}_{t-1}} \leftarrow f(\nabla_{\mathcal{W}_{t-1}})$
   \STATE $\mathcal{W}^s_t \leftarrow \mathcal{W}^{s-1}_{t-1} + \Delta_{s-1,\mathcal{W}_{t-1}}$
   \STATE  $L_{cons} \leftarrow  \|\mathcal{W}^S_t -\mathcal{W}_0 \|^2$
   \STATE $L_{e,i} \leftarrow -\log{\theta_{\mathcal{W}_t}(y^e_i \mid x^e_i)}$
    \STATE $L_i \leftarrow L_{e,i} + \eta L_{cons}$
    \STATE Back-propagate $L_i$
   \STATE Update $f$
    \ENDFOR
    \ENDFOR
   \STATE \textbf{return} $f$
   \end{algorithmic}
\end{algorithm}
\subsection{Lightweight Training} \label{sec:Lightweight Training}
As shown in Section \ref{sec:Meta Learning Based Model Editing}, previous MLME methods typically use the weighted sum of the generalization loss $ L_e $ and the specificity loss $ L_{loc} $ as the meta-loss for training the hypernetwork. While this meta-loss effectively balances the trade-off between editing new knowledge and preserving original knowledge, we show in Section \ref{sec:Training Time Analysis of MLME} that computing the loss $ L_{loc} $ leads to inefficient training. To improve training efficiency while preserving the original model's knowledge as much as possible, inspired by RLEdit, we replace the specificity loss with an $ l_2 $-norm constraint loss $L_{cons}$ on the weight updates. Therefore, the final meta-loss of EMSEdit is formulated as:
\begin{equation}
\small
   L_{meta} = L_e + \eta L_{cons} = L_e + \eta \mathop{\sum}_{s=0}^{S} \|\Delta_{s,\mathcal{W}_t} \|^2
\end{equation}
\begin{table*}[ht]
\centering
\resizebox{1\textwidth}{!}{
\begin{tabular}{c|rrrrrr|rrrrrr|rrrrrr}
\toprule
\multirow{3}{*}{\textbf{Method}}  
& \multicolumn{6}{c|}{\textbf{GPT-J (6B)}}  
& \multicolumn{6}{c|}{\textbf{LLaMA-3 (8B)}}  
& \multicolumn{6}{c}{\textbf{Gemma-2 (9B)}} \\
\cmidrule(lr){2-7} \cmidrule(lr){8-13} \cmidrule(lr){14-19}
& \multirow{2}{*}{\textbf{Eff.}$_\uparrow$}  
& \multirow{2}{*}{\textbf{Gen.}$_\uparrow$}  
& \multirow{2}{*}{\textbf{Spe.}$_\uparrow$}  
& \multirow{2}{*}{\textbf{Avg.}$_\uparrow$}  
& \multicolumn{2}{c|}{\textbf{Time (s) / Edit}}  
& \multirow{2}{*}{\textbf{Eff.}$_\uparrow$}  
& \multirow{2}{*}{\textbf{Gen.}$_\uparrow$}  
& \multirow{2}{*}{\textbf{Spe.}$_\uparrow$}  
& \multirow{2}{*}{\textbf{Avg.}$_\uparrow$}  
& \multicolumn{2}{c|}{\textbf{Time (s) / Edit}}  
& \multirow{2}{*}{\textbf{Eff.}$_\uparrow$}  
& \multirow{2}{*}{\textbf{Gen.}$_\uparrow$}  
& \multirow{2}{*}{\textbf{Spe.}$_\uparrow$}  
& \multirow{2}{*}{\textbf{Avg.}$_\uparrow$}  
& \multicolumn{2}{c}{\textbf{Time (s) / Edit}} \\
\cmidrule(lr){6-7} \cmidrule(lr){12-13} \cmidrule(lr){18-19}
&   &   &   &   & \textbf{Train$_\downarrow$}  & \textbf{Val$_\downarrow$}  
&   &   &   &   & \textbf{Train$_\downarrow$}  & \textbf{Val$_\downarrow$}  
&   &   &   &   & \textbf{Train$_\downarrow$}  & \textbf{Val$_\downarrow$} \\
\midrule
\multicolumn{19}{c}{\textbf{ZsRE-Sequential Batch Editing-30$\times$600}} \\
\midrule
\textbf{FT} &50.79&47.98&29.47 &42.75&-&0.14&69.67&62.27&32.22&54.72&-&0.16&7.06&6.38&9.96&7.80&-&0.28 \\
\textbf{MEND} &0.60&0.62&0.52&0.58&\textbf{0.10}&\textbf{0.04}&0.21&0.15&0.06&0.14&0.15&0.06&21.43&19.88&14.05&18.45&\textbf{0.10}&\textbf{0.07} \\
\textbf{RLEdit} & 83.76 & 81.47 & 29.13&64.79 & 0.50 &0.05 & 90.16 & 88.78 & 42.04&73.66 & 0.67 & 0.09 & 84.60 & 79.66 & 35.67&66.64 & 0.67 & 0.11 \\
\textbf{UltraEdit} &65.53  &61.59 & \textbf{31.34}&52.82 &- & \textbf{0.04}& 87.41 &84.29  &\textbf{49.94}&73.88  &-  &0.06 &83.26 &77.17  &\textbf{50.03}&70.15 & -& 0.08 \\
\textbf{EMSEdit} & \textbf{93.61} & \textbf{90.34} & 30.64 & \textbf{71.53} &{0.17}& 0.16 & \textbf{95.40} & \textbf{93.91} & {42.18} &\textbf{77.16}& \textbf{0.19} & 0.11 &\textbf{89.55} & \textbf{85.11} & 36.15 &\textbf{70.27}& 0.25 & 0.16 \\
\midrule
\multicolumn{19}{c}{\textbf{CounterFact-Sequential Batch Editing-20$\times$400}} \\
\midrule
\textbf{FT} &40.72&33.72&\textbf{62.29}&45.58&-&0.16&83.70&52.44&\textbf{51.63}&62.59&-&0.18&48.39&36.27&51.24&45.30&-&0.31 \\
\textbf{MEND} &42.12&42.05&57.88&47.35&\textbf{0.08}&\textbf{0.08}&48.67&49.44&50.32&49.48&\textbf{0.14}&\textbf{0.07}&25.14&26.23&\textbf{72.70}&41.36&\textbf{0.22}&0.11 \\
\textbf{RLEdit} & 61.34 & 45.76 & {54.06}&53.72 & 0.53 & 0.15 & 63.95 & 57.56 & 41.72 &54.41& 0.54 & \textbf{0.07} & 60.31 & 50.16 & 48.34&52.94 & 0.65 & 0.08 \\
\textbf{UltraEdit} &  43.23  & 32.81 & 58.85 &44.96& -& 0.06   &63.55  & 48.37 & 43.87&51.93&- & 0.06 & 66.43&49.97 & 49.97&55.46&-&\textbf{0.07} \\
\textbf{EMSEdit} & \textbf{94.43} & \textbf{65.56} & 44.48&\textbf{66.82} & {0.26} & {0.12} & \textbf{94.24} &\textbf{ 79.49} & 40.56 &\textbf{71.43}& {0.30} & 0.14 & \textbf{89.74} & \textbf{70.12} & 42.37&\textbf{67.41} & 0.28 & 0.16\\
\midrule
\multicolumn{19}{c}{\textbf{ZsRE-Batch Editing-16384}} \\
\midrule
\textbf{FT} &26.90&26.20&\textbf{27.31}&26.80&-&0.14&38.35&36.77&32.11&35.74&-&0.16&3.21&3.01&6.37&4.20&-&0.27 \\
\textbf{MEND} &0.01&0.01&0.00&0.01&0.07&\textbf{0.02}&0.01&0.02&0.02&0.02&0.11&\textbf{0.04}&23.58&22.68&13.18&19.81&0.15&0.06 \\
\textbf{MALMEN} &46.05&44.14&27.29&39.16&\textbf{0.05}&\textbf{0.02}&69.28&66.59&45.46&60.44&\textbf{0.07}&\textbf{0.04}&78.76&71.85&\textbf{42.18}&64.26&\textbf{0.13}&\textbf{0.05} \\
\textbf{EMSEdit} &\textbf{82.66}&\textbf{72.73}&25.81&\textbf{60.40}&0.07&0.04&\textbf{84.15}&\textbf{80.94}&\textbf{49.41}&\textbf{71.50}&0.11&0.07&\textbf{96.57} &\textbf{90.34}&39.31&\textbf{75.41}&0.19&0.09 \\
\midrule
\multicolumn{19}{c}{\textbf{CounterFact-Batch Editing-8192}} \\
\midrule
\textbf{FT} &15.17&17.66&83.22&38.68&-&0.14&7.07&9.63&\textbf{89.37}&35.36&-&0.15&49.68&31.12&49.31&43.37&-&0.28 \\
\textbf{MEND} &44.76&46.60&54.53&48.63&0.09&0.03&43.73&45.11&55.80&48.21&0.10&0.03&15.03&19.28&\textbf{80.80}&38.37&0.13&0.04 \\
\textbf{MALMEN}&66.92&29.64&\textbf{75.86}&57.47&\textbf{0.04}&\textbf{0.03}&51.06&29.63&78.42&53.04&\textbf{0.06}&\textbf{0.03}&94.57&58.86&62.46&71.96&\textbf{0.09}&\textbf{0.04} \\
\textbf{EMSEdit} &\textbf{99.82}&\textbf{50.07}&61.44&\textbf{70.44}&0.05&0.04&\textbf{97.96}&\textbf{60.42}&57.83&\textbf{72.07}&0.08&0.05& \textbf{99.06}&\textbf{73.24}&50.03&\textbf{74.11}&0.10&0.06 \\
\bottomrule
\end{tabular}}
\caption{Comparison of Editing Methods across GPT-J (6B), Llama-3 (8B), and Gemma2 (9B) on ZsRE and CounterFact Datasets in Sequential Editing Task. Eff.: Efficacy, Gen.: Generalization, Spe.: Specificity. A setting of $30\times600$ indicates performing 600 sequential batch edits, each with a batch size of 30. The number following batch editing denotes the batch size.
}
\label{tab:editing_comparison}
\vspace{-0.4cm}
\end{table*}

\subsection{Step-specific Hypernetwork}
EMSEdit achieves a multi-step optimization process similar to traditional deep learning training through MSBP. In sequential batch editing, the hypernetwork must model both \textbf{inter-step variance} (across BP steps) and \textbf{inter-batch variance} (across batches), which is challenging for a single small hypernetwork. To address this, we adopt step-specific hypernetworks that specialize in capturing inter-batch differences. Specifically, EMSEdit performs editing over $S$ BP steps, where the $s$-th step is handled by a hypernetwork $f_s$. These hypernetworks form a set $\mathcal{F}={f_1,\dots,f_S}$. The corresponding training procedure is shown in Algorithm~\ref{alg:smtrain}.

Each hypernetwork shares the same architecture as prior MLME methods~\cite{mitchell2022fast,tan2024massiveeditinglargelanguage}. As gradients typically diminish over steps, learning the gradient-to-pseudo-gradient transformation becomes easier in later stages. Accordingly, we apply a rank decay strategy to reduce the editing burden, defining the rank at step $s$ as $r_s = r \mid s$, where $r$ is a predefined rank.
\subsection{Step-wise Hypernetwork Updating}
Batch editing performs a single edit with a large batch size, where the primary source of variance arises from inter-step variance. In this setting, a single hypernetwork is sufficient and reduces memory overhead compared to maintaining multiple networks. Accordingly, batch editing training employs a single hypernetwork $f$ to learn multi-step gradient transformations, as it better captures underlying transformation patterns. To further reduce memory consumption caused by gradient accumulation, especially with large batch sizes, we adopt step-wise hypernetwork updates, updating the hypernetwork at each backpropagation step. The pseudo-code of EMSEdit hypernetwork training for batch editing is shown in Algorithm~\ref{alg:smtrain-batch}.
\section{Experiments}\label{sec:exp}
\subsection{Experimental Settings}\label{sec:exp:setting}
% LLM 、 DATASETS 、METRICS、BASELINES
\textbf{LLMs:} GPT-J (6B) \cite{gpt-j}, LLaMA-3 (8B) \cite{llama3modelcard}, and Gemma-2 (9B) \cite{team2024gemma}.

\textbf{DataSets \& Metrics} We consider two common datesets in model editing: CounterFact \cite{Meng2022Locating} and ZsRE \cite{zsre}. In line with prior work \cite{li2025reinforced,tan2024massiveeditinglargelanguage}, we evaluate editing performance using three metrics: Efficacy, Generalization, and Specificity.  Details are shown in Appendix \ref{appendix:Experimental detials}.

\textbf{Baselines} Our work primarily addresses the limitations of MLME. For sequential editing, the baselines include FT \cite{zhang2024comprehensive}, MEND \cite{mitchell2022fast}, RLEdit \cite{li2025reinforced}, and UltraEdit \cite{gu2025ultraedit}. For batch editing, we use FT \cite{zhang2024comprehensive}, MEND \cite{mitchell2022fast}, and MALMEN \cite{tan2024massiveeditinglargelanguage} as our baselines. 

\textbf{Experimental Details} These details are shown in Appendix \ref{appendix:Experimental detials}.
\subsection{Main Results}\label{sec:main_res}
\subsubsection{Resuts on Sequential Batch Editing}
We evaluate EMSEdit on sequential batch editing using GPT-J (6B), LLaMA-3 (8B), and Gemma-2 (9B) on the ZsRE and CounterFact datasets. As shown in the upper part of Table~\ref{tab:editing_comparison}, EMSEdit achieves the best overall performance across all models and datasets, with substantial gains in Efficacy and Generalization over existing MLME methods such as RLEdit and UltraEdit. For example, on ZsRE, EMSEdit reaches 93.61/95.40/89.55 in Efficacy, outperforming RLEdit by a large margin. Although its Specificity is slightly lower than UltraEdit in some cases, EMSEdit attains the highest Average score, indicating a strong balance between editing effectiveness and model stability. These results demonstrate that MSBP enables effective learning from limited editing data.

In terms of efficiency, EMSEdit maintains competitive editing speed, with training and validation time per edit comparable to or lower than other MLME methods and significantly faster than fine-tuning. Overall, EMSEdit achieves a superior trade-off between performance and efficiency under sequential editing.
\subsubsection{Resuts on Batch Editing}
In the batch editing scenarios on ZsRE and CounterFact, EMSEdit continues to deliver the strongest performance among all baselines.
Across all three models, EMSEdit achieves the highest scores in Efficacy, Generalization, and Average, demonstrating remarkable robustness under large-scale editing.
For example, on ZsRE (Gemma-2), EMSEdit attains an Average of 75.41, surpassing MALMEN (64.26) by a significant margin.
Similarly, on CounterFact, EMSEdit obtains an Average of 70.44, 72.07, and 74.11 on GPT-J, LLaMA-3, and Gemma-2, respectively—establishing new state-of-the-art results in batch editing.
Although EMSEdit’s Specificity is slightly below MALMEN in isolated cases, its overall balance across metrics is more consistent, suggesting stable model behavior even when multiple edits are performed simultaneously.

From an efficiency perspective, EMSEdit remains lightweight and scalable.
Its per-edit training and validation time (e.g., 0.07 s / 0.04 s on GPT-J for ZsRE) is on par with MALMEN while achieving much higher performance, showing that EMSEdit’s design yields both computational efficiency and superior editing quality.
\begin{figure}[t]
    \centering
    \includegraphics[trim=5 5 5 5, clip, width=1\linewidth]{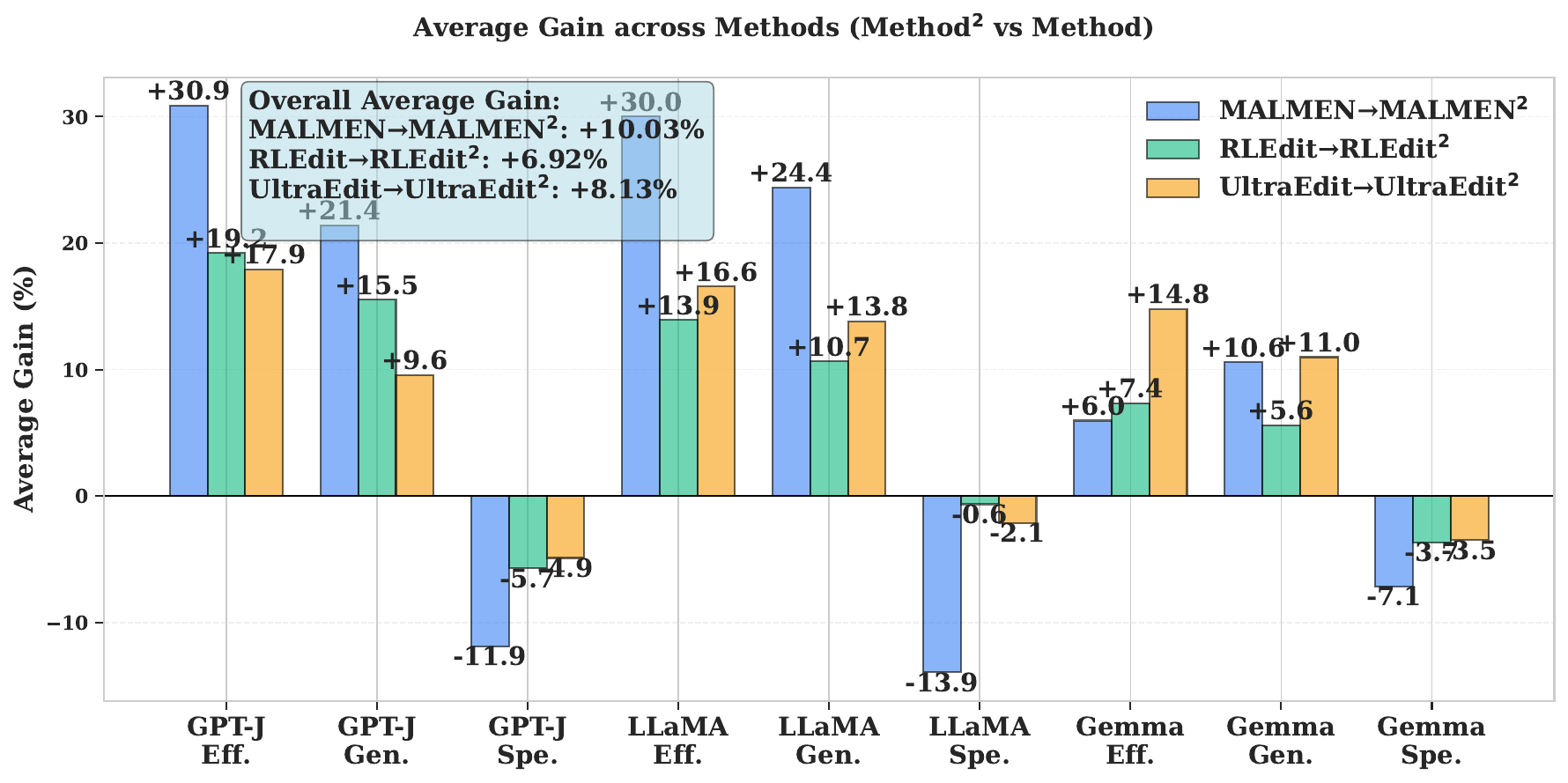}
    \caption{Performance Gains of MSBP over Existing Methods.}
    \label{fig:gain-res}
\end{figure}
\subsubsection{Performance Gains of MSBP over Existing Methods}\label{sec:performance_gain}
To evaluate the effectiveness of MSBP, we apply a two-step BP mechanism to existing MLME methods, including MALMEN, RLEdit, and UltraEdit, and evaluate them on sequential batch and batch editing tasks. Figure~\ref{fig:gain-res} shows the average performance gains across different LLMs, where MALMEN$^2$, RLEdit$^2$, and UltraEdit$^2$ denote the enhanced versions. MSBP consistently improves Efficacy and Generalization (e.g., +20.55\% for MALMEN), while causing moderate declines in Specificity (e.g., -3.3\% for RLEdit), indicating increased interference with unedited knowledge due to overfitting. Overall, MSBP yields clear average performance gains of 10.03\%, 6.92\%, and 8.13\% for MALMEN, RLEdit, and UltraEdit, respectively, demonstrating its effectiveness across methods and models. Additional results under different sequential and batch settings are provided in Appendix~\ref{sec:performance_varies}.
\begin{figure}[t]
    \centering
    \includegraphics[trim=5 5 5 5, clip, width=1\linewidth]{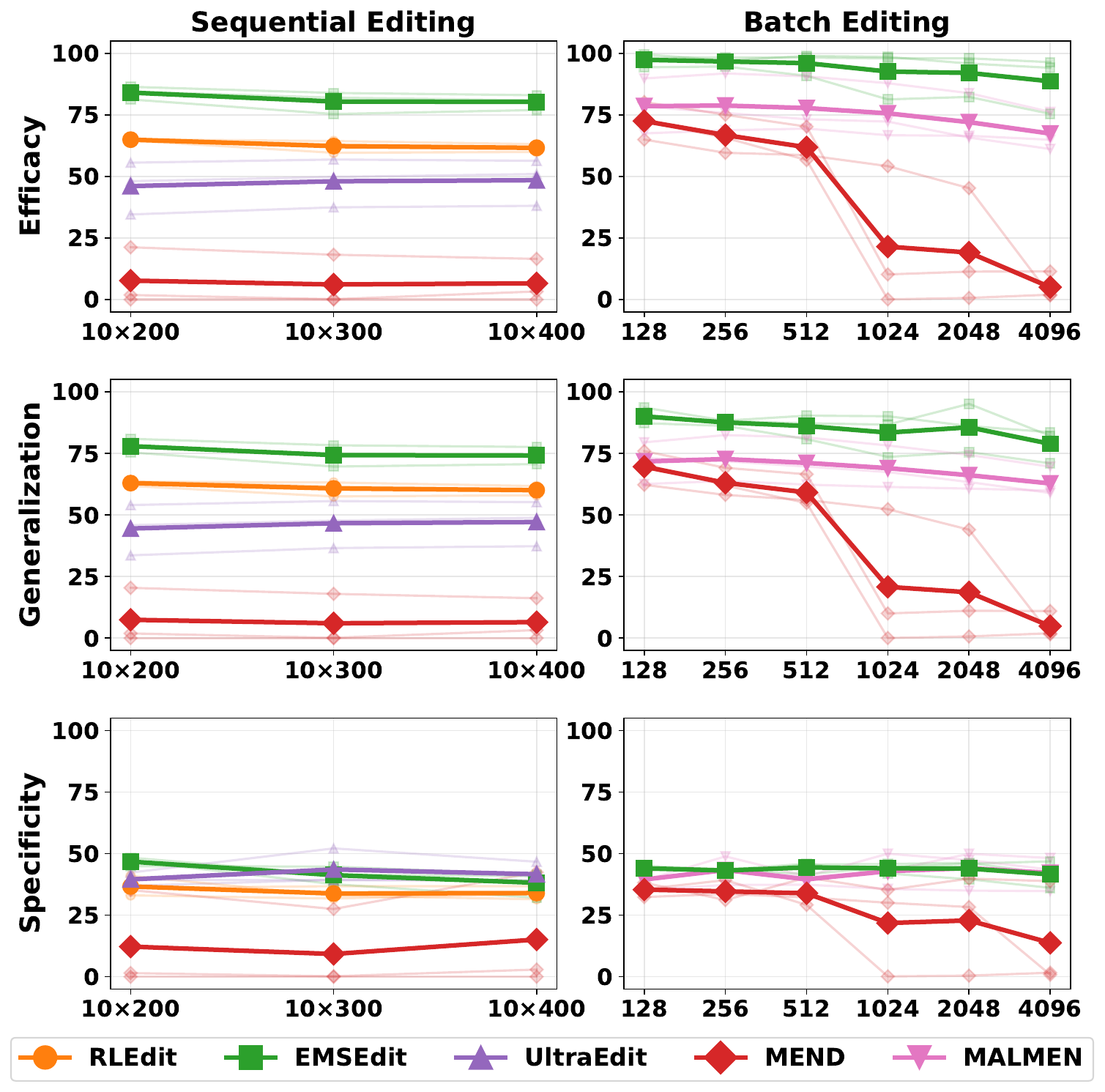}
    \caption{Average Editing Performance Across LLMs. Individual method performance is shown in lighter shades.}
    \label{fig:ripple}
\end{figure}
\subsubsection{Comparison Results on multi-hop reasoning dataset}\label{sec:multi-hop}
To evaluate the performance of EMSEdit on multi-hop reasoning model editing, we conduct experiments on the RippleEdit \cite{cohen2024evaluating} dataset using LLaMA-3 (8B), GPT-J (6B), and Gemma-2 (9B). We follow \cite{zhang-etal-2024-dafnet} to split the dataset into training and test sets, and maintain the same experimental settings as in the main experiments. We report the results in Figure~\ref{fig:ripple}. As shown, EMSEdit consistently demonstrates superior performance in both sequential and batch editing scenarios across different models. It not only achieves higher editing efficacy and generalization but also maintains strong specificity, confirming its robustness and scalability.

\subsection{Ablation Study}
We verify the number of BP steps selected by EMSEdit, as well as the effectiveness of \textit{Step-specific Hypernetwork} (corresponding to \textit{identical hypernetwork}) and \textit{rank decay} (corresponding to \textit{consistent rank}) for sequential editing, and \textit{Step-wise Hypernetwork Updating} (corresponding to \textit{No step-wise}) and \textit{Identical Hypernetwork} (corresponding to \textit{diff hypernetwork}) for batch editing. We conduct 8192-sized batch editing and 30$\times$400 sequential batch editing on the LLaMA-3 (8B) model using the ZsRE dataset. The results are shown in Table \ref{tab:ablation}, with EMSEdit results highlighted in cyan.

\begin{table}[t]
  \centering
  \caption{Ablation Study.}
           \vspace{-0.3cm}
    \resizebox{0.48\textwidth}{!}{\begin{tabular}{lrrrrr}
    \toprule
    \multicolumn{6}{c}{\textbf{Bath Size: 8192}} \\
    \midrule
    \textbf{Metrics} & \multicolumn{1}{l}{\textbf{Eff.}} & \multicolumn{1}{l}{\textbf{Gen.}} & \multicolumn{1}{l}{\textbf{Spe.}} & \multicolumn{1}{l}{\textbf{Train (s) / Edit}} & \multicolumn{1}{l}{\textbf{Val (s) / Edit}} \\
    \midrule
    \textbf{BP Step 1} & 70.18 & 67.22 & 46.01 & 0.05  & 0.04 \\
    \midrule
    \rowcolor[HTML]{71c8bf}\textbf{BP Step 2 (EMSEdit)} & 85.91 & 80.64 & 44.44 & 0.11  & 0.07 \\
    \midrule
    \textbf{BP Step 3} & 91.19 & 84.50  & 47.60  & 0.15  & 0.11 \\
    \midrule
    \textbf{BP Step 4} & 94.55  & 86.91  & 49.01  & 0.20   & 0.14 \\
    \midrule
    \textbf{Diff HyperNetwork} & 83.57 & 78.13 & 46.88 & 0.12  & 0.08 \\
    \midrule
    \textbf{No Step-wise} & \multicolumn{5}{c}{Out of Memory} \\
    \bottomrule
    \toprule
    \multicolumn{6}{c}{\textbf{Sequence Length$\times$Bath Size: 30$\times$400}} \\
    \midrule
    \textbf{BP Step 1} & 89.31 & 87.57 & 43.47 & 0.11  & 0.06 \\
    \midrule
    \rowcolor[HTML]{71c8bf}\textbf{BP Step 2 (EMSEdit)} & 95.61 & 94.09 & 44.94 & 0.18  & 0.11 \\
    \midrule
    \textbf{BP Step 3} & 88.08 & 85.86 & 36.88 & 0.29  & 0.19 \\
    \midrule
    \textbf{BP Step 4} & 55.19 & 49.39 & 16.42 & 0.56  & 0.21 \\
    \midrule
    \textbf{Identical HyperNetwork} & 93.57 & 91.83 & 38.66 & 0.35  & 0.12 \\
    \midrule
    \textbf{Consistent Rank} & 94.52 & 92.53 & 44.02 & 0.20  & 0.12 \\
    \bottomrule
    \end{tabular}}%
  \label{tab:ablation}%
           \vspace{-0.4cm}
\end{table}

Regarding the number of BP steps, although a BP step of 3 and 4 achieves better editing performance in batch editing, its training and validation time nearly doubles. Choosing 2 steps achieves a better balance between performance and efficiency. For sequential batch editing, using 2 BP steps yields the best balance between performance and efficiency.

In terms of the hypernetwork design, the results indicate that using different hypernetworks in batch editing and the same hypernetwork in sequential batch editing both lead to worse editing performance. This may be due to the differing nature of batch and sequential editing tasks. Additionally, maintaining a consistent hypernetwork rank in sequential batch editing does not yield better editing results. Furthermore, in batch editing, not using step-wise updating requires substantial memory for gradient accumulation—we also experimented with a batch size of 1024 and still encountered out-of-memory errors.
\begin{figure}[htbp]
        \centering
        \includegraphics[trim=5 5 5 5, clip, width=1\linewidth]{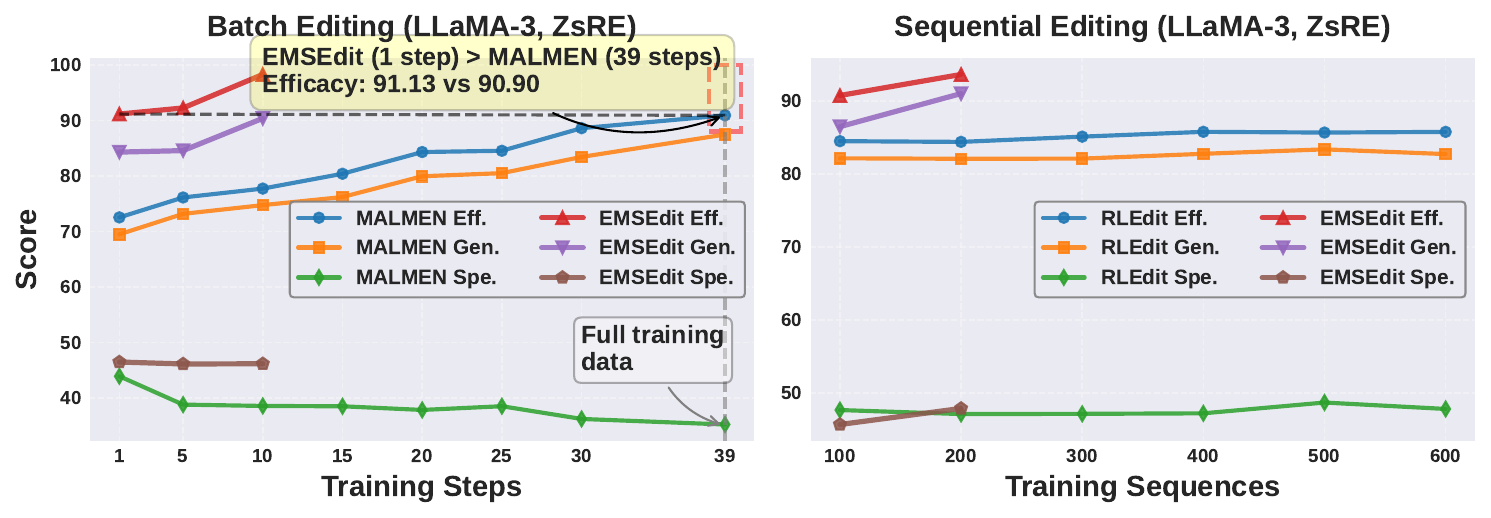}
    \caption{Data Efficiency of EMSEdit.}
    \label{fig:data_efficiency}
\end{figure}
\subsection{Data Efficiency of EMSEdit}
We conduct sequential and batch editing experiments on ZsRE with LLaMA-3 to evaluate the data efficiency of EMSEdit. For sequential editing, we use a batch size of 4 and vary the number of training sequences from 100 to 600, with 600 validation sequences. For batch editing, we fix the validation batch size to 4096 and vary the training steps from 1 to 39 (full data). Other settings follow the main experiments, and results are shown in Fig.~\ref{fig:data_efficiency}.

In batch editing, MALMEN’s performance improves with increasing training steps and reaches its best performance only after all training data have been used. In contrast, EMSEdit surpasses MALMEN’s best performance with only a single training step and continues to improve with additional steps. In sequential editing, RLEdit shows a modest performance gain as the training sequence length increases, but plateaus at a length of 400 with a maximum efficacy of 85.80. By comparison, EMSEdit exceeds RLEdit’s best performance with a sequence length of just 100. These results demonstrate that EMSEdit is substantially more data-efficient than both MALMEN and RLEdit, achieving superior performance with far less data, even when the baselines are trained on much larger datasets. This suggests that MSBP improves the utilization efficiency of editing data, analogous to conventional deep learning training where repeated iterations over the same dataset enhance learning effectiveness. Consistent with this interpretation, Appendix~\ref{appendix:loss-vary} shows that under MSBP, the loss in EMSEdit progressively decreases as the number of backpropagation steps increases.
\subsection{Memory Efficiency of EMSEdit}
\begin{figure}[t]
    \centering
    \includegraphics[trim=5 5 5 5, clip, width=1\linewidth]{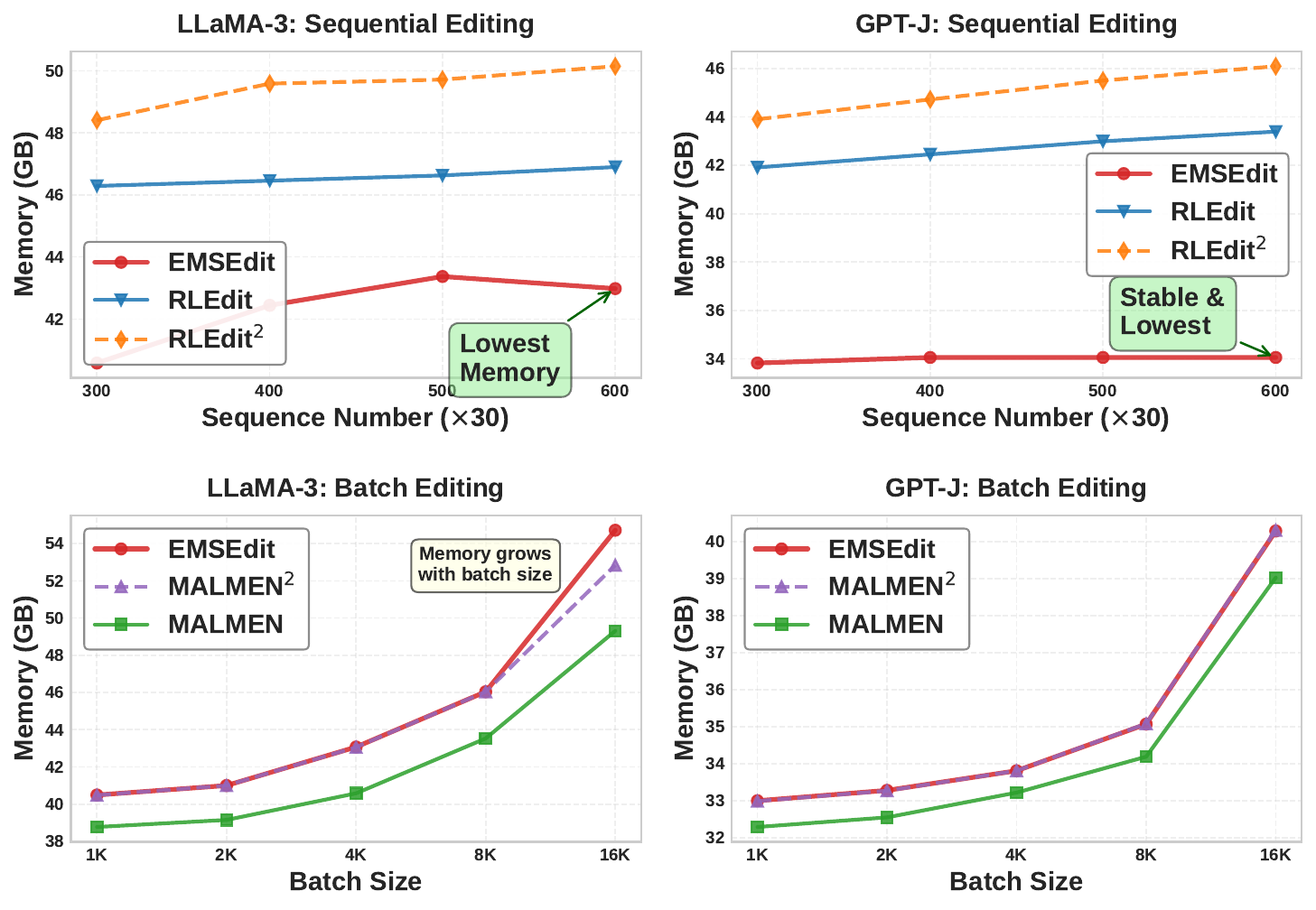}
    \caption{Memory Efficiency of EMSEdit.}
    \label{fig:memory_effiency}
\end{figure}
Figure~\ref{fig:memory_effiency} reports the memory usage of EMSEdit, MALMEN, and RLEdit in our main experiments. The results show that EMSEdit exhibits high memory efficiency, particularly in sequential editing, where it outperforms MALMEN and RLEdit by a wide margin. In batch editing, EMSEdit and MALMEN$^2$ display nearly identical memory profiles, indicating comparable efficiency, although EMSEdit incurs slightly higher memory usage at larger batch sizes. Overall, EMSEdit demonstrates good scalability and stable memory behavior across all settings, making it a practical choice for memory-constrained or resource-limited scenarios.
\section{Conlusion}
We propose EMSEdit, a new MLME method that improves data and training efficiency. By leveraging multi-step backpropagation (MSBP), EMSEdit performs effective editing in low-data regimes, while replacing KL divergence with $l_2$ regularization reduces computational overhead without degrading performance. Our step-specific and step-wise hypernetwork designs further improve efficiency and performance across sequential and batch settings. Experiments on GPT-J, LLaMA-3, and Gemma-2 over ZsRE and CounterFact show that EMSEdit achieves state-of-the-art results and demonstrate the broader potential of MSBP for model editing.
\section*{Acknowledgements}
This work was supported by the National Key Research and Development Program of China under Grant 2024YFB4506200, the
Science and Technology Innovation Program of Hunan Province
under Grant 2024RC1048, the National Key Laboratory Foundation Project under Grant 2024-KJWPDL-14, the National Natural Science Foundation of China under Grant 62402506, and the Innovation Reserch Foundation of (NUDT) (HQKYZH2025KD004).
\bibliographystyle{ACM-Reference-Format}
\bibliography{References}
\balance

%%%%%%%%%%%%%%%%%%%%%%%%%%%%%%%%%%%%%%%%%%%%%%%%%%%%%%%%%%%%
\appendix
\begin{figure*}[t]
    \centering
    \includegraphics[width=1\linewidth]{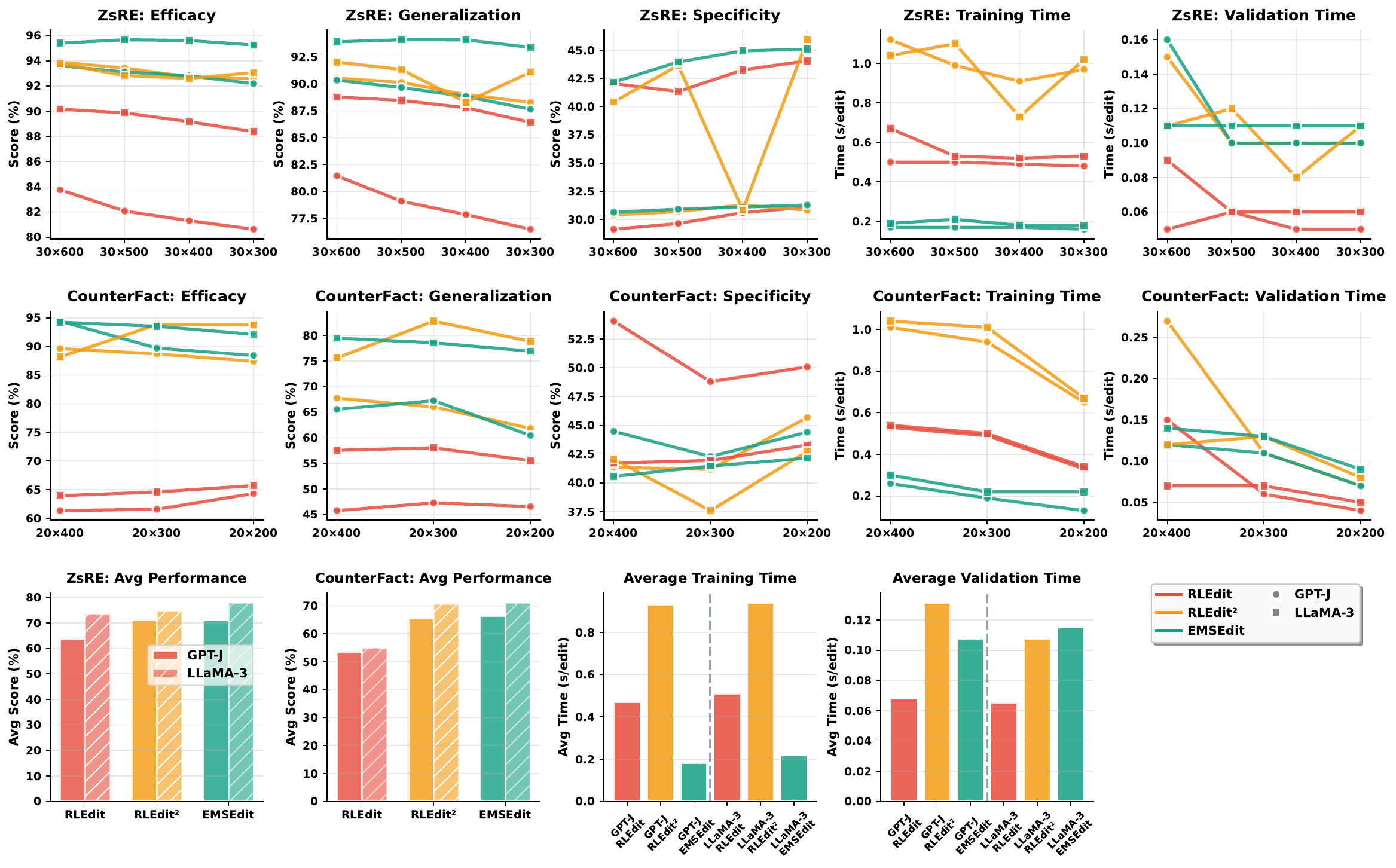}
    \caption{Editing performance of EMSEdit and baselines with respect to the number of editing sequences.}
    \label{fig:supp_seq_res}
\end{figure*}
\section{Editing Performance vs. Number of Sequences/Batches}\label{sec:performance_varies}
\begin{figure}[h]
    \centering
    \includegraphics[width=1\linewidth]{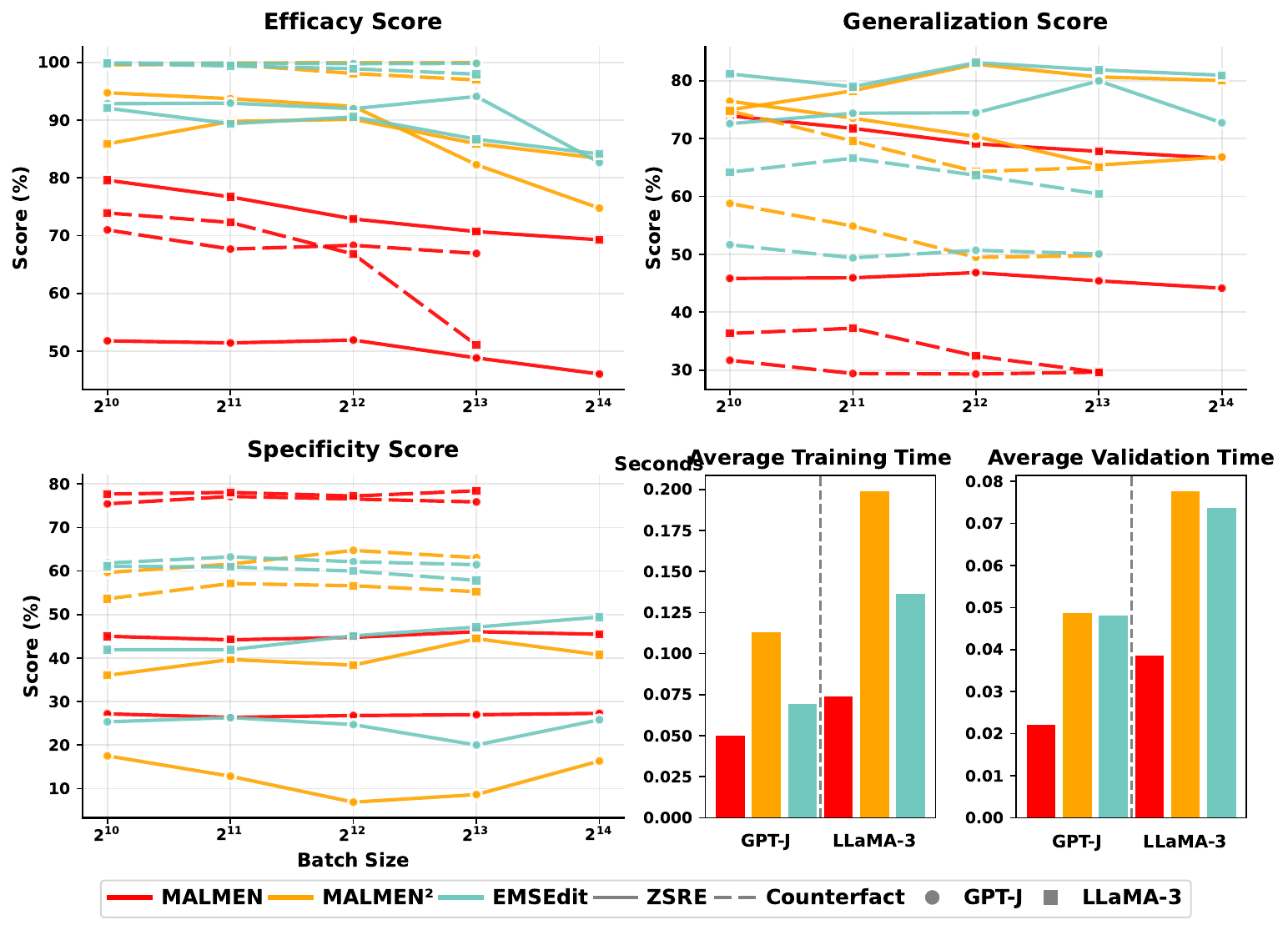}
    \caption{Editing performance of EMSEdit and baselines with respect to the number of editing batches.}
    \label{fig:supp_batch_res}
\end{figure}

We conduct experiments with RLEdit, RLEdit$^2$, MALMEN, MALMEN$^2$, and EMSEdit under both sequential and batch editing settings. The results are presented in Figures \ref{fig:supp_seq_res} and \ref{fig:supp_batch_res}, respectively.

In sequential editing, RLEdit$^2$ and EMSEdit achieve comparable editing performance across different datasets, models, and editing setups, both significantly outperforming RLEdit, except in terms of Specificity on the CounterFact dataset. Notably, EMSEdit demonstrates slightly better average performance than RLEdit$^2$. In terms of efficiency, RLEdit achieves the highest validation efficiency, which is intuitive since it performs editing using only single-step updates, whereas EMSEdit and RLEdit$^2$ rely on two-step updates. Regarding training efficiency, EMSEdit performs the best, followed by RLEdit, while RLEdit$^2$ shows the lowest efficiency. The batch editing results are consistent with those of sequential editing. EMSEdit and MALMEN$^2$ achieve comparable overall performance; however, both MALMEN$^2$ and EMSEdit show a noticeable drop in Specificity on the CounterFact dataset compared with MALMEN.

Across both sequential and batch settings, the results suggest that multi-step backpropagation (MSBP) improves Efficacy and Generalization but degrades Specificity. This degradation likely arises from overfitting to the injected knowledge. We leave mitigating this issue to future work.
\section{Analysis of the Loss Variation}\label{appendix:loss-vary}
\begin{figure}[h]
    \centering
    \includegraphics[width=1\linewidth]{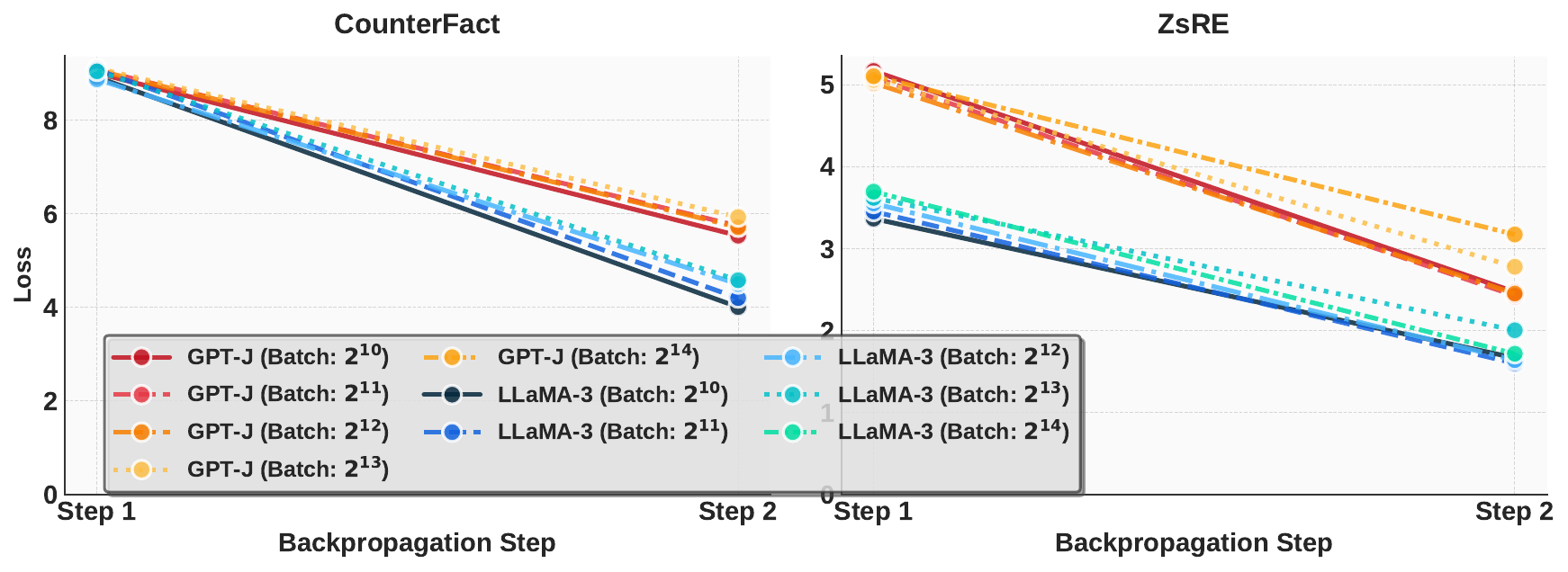}
    \caption{The variation of loss across different BP steps for different batch edit sizes.}
    \label{fig:loss2}
\end{figure}

In this part, we analyze the trend of training loss across different BP steps in EMSEdit's MSBP process. Since the differences in loss across BP steps are more pronounced in the batch editing experiments, we focus our analysis on this setting. Specifically, we record the loss at each BP step in the batch editing experiment. We show the loss trend in Figure \ref{fig:loss2}. It can be observed that for both datasets and models, the loss at BP Step 2 is consistently lower than at BP Step 1. This indicates that the introduction of MSBP in MLME has a similar effect to multi-epoch BP in standard deep learning training, where the loss gradually decreases with more iterations.
\begin{figure}[h]
    \centering
    \includegraphics[width=0.99\linewidth]{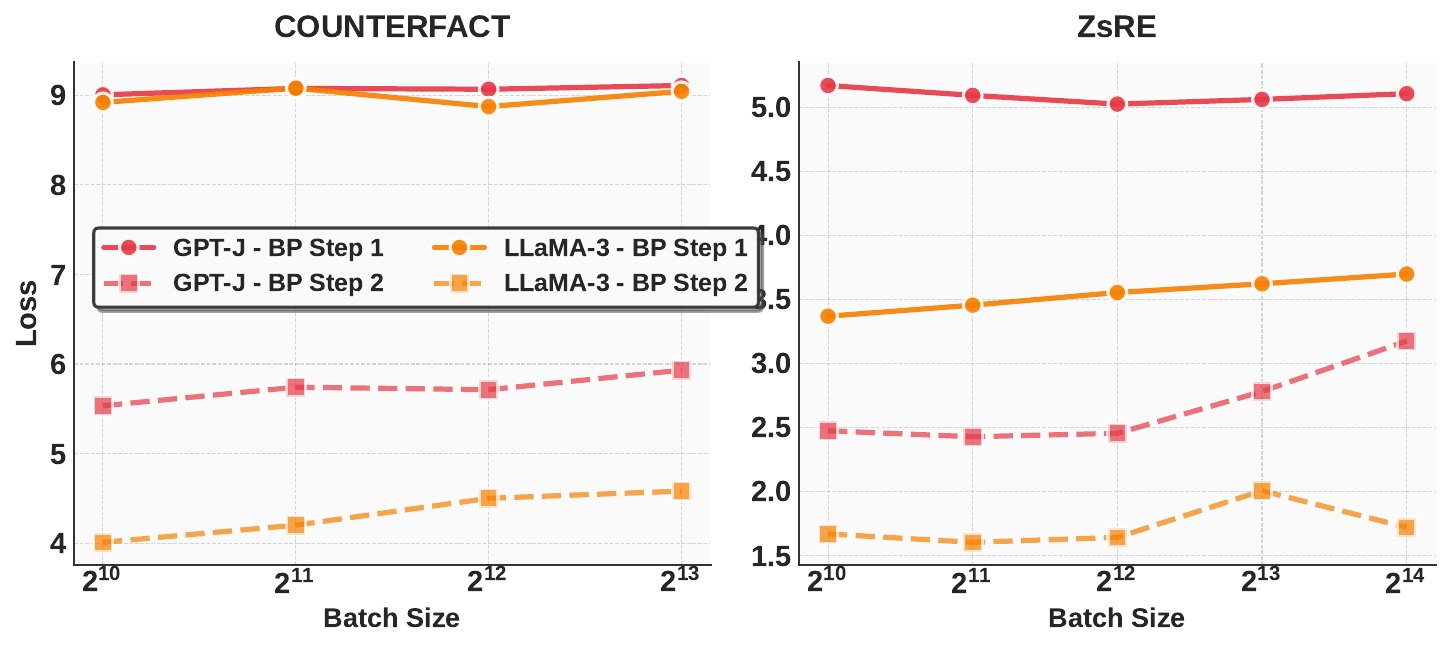}
    \caption{The loss at different BP steps varies with the number of batch edits.}
    \label{fig:loss1}
\end{figure}
We also present the variation of the loss norm across different BP steps with respect to the editing batch size. As shown in Figure \ref{fig:loss1}, it can also be observed that for both datasets and models, the loss at BP Step 2 is consistently lower than at BP Step 1. Moreover, the overall increase in loss with larger batch sizes aligns with intuition. This indicates that the introduction of MSBP in MLME has a similar effect to multi-step BP in standard deep learning training, where the loss gradually decreases with more iterations.

\begin{figure}[h]
    \centering
    \includegraphics[width=0.8\linewidth]{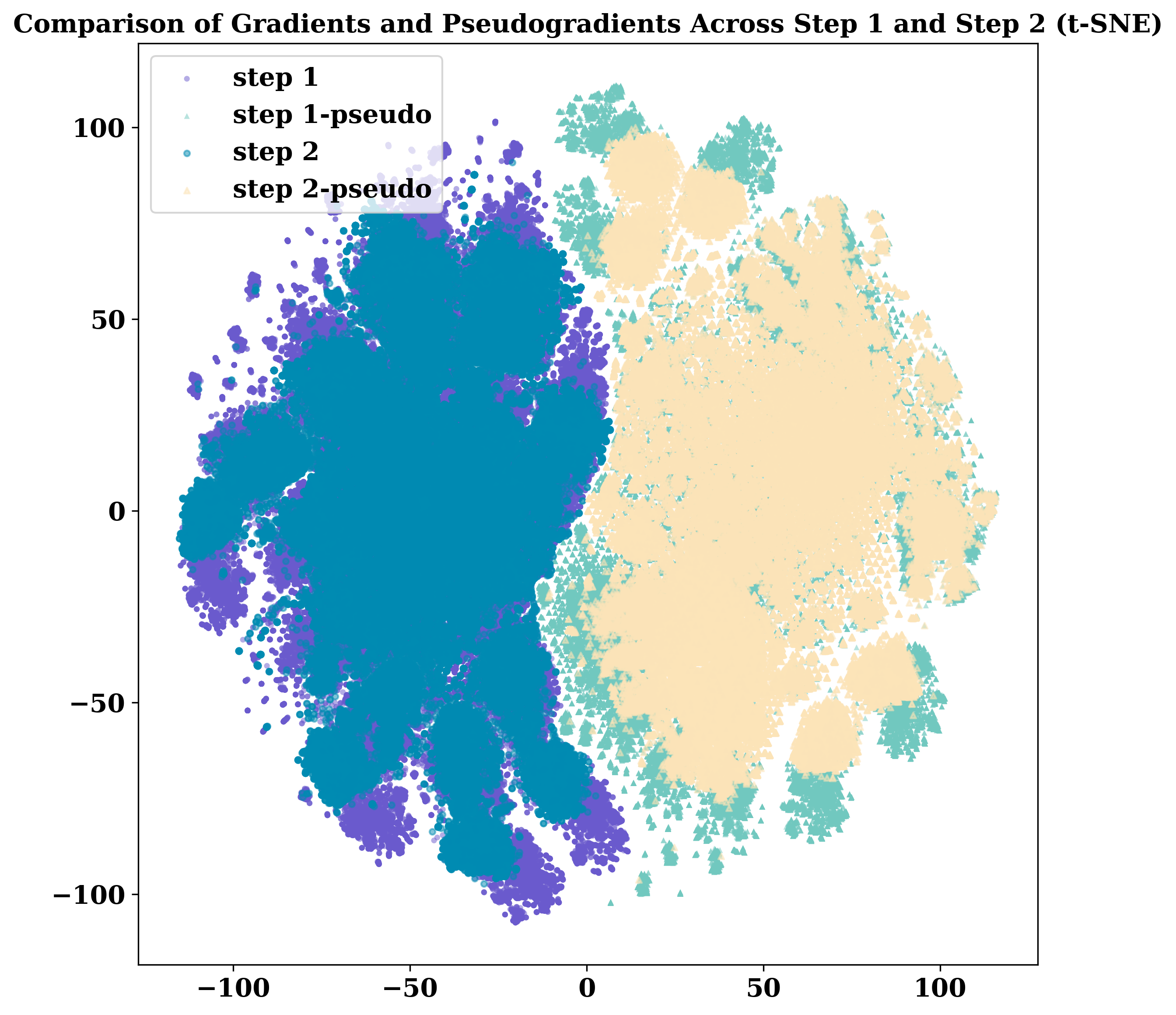}
    \caption{t-SNE visualization of gradients and pseudo-gradients after dimensionality reduction.}
    \label{fig:loss_visual}
\end{figure}

Additionally, we visualize the gradients and pseudo-gradients at different BP steps during the validation of $2^{13}$ batch edits, as shown in Figure \ref{fig:loss_visual}. It can be observed that there are clear differences between the gradients and pseudo-gradients at both Step 0 and Step 1, which are attributed to the transformation effect of the trained hypernetwork. On the other hand, the distributions of gradients and pseudo-gradients at Step 1 are more dispersed compared to those at Step 2. This indicates that MSBP reduce the spread of the gradient distribution, which is consistent with our previous analysis of the loss across different BP steps.
\begin{figure}[t]
    \centering
    \includegraphics[width=0.99\linewidth]{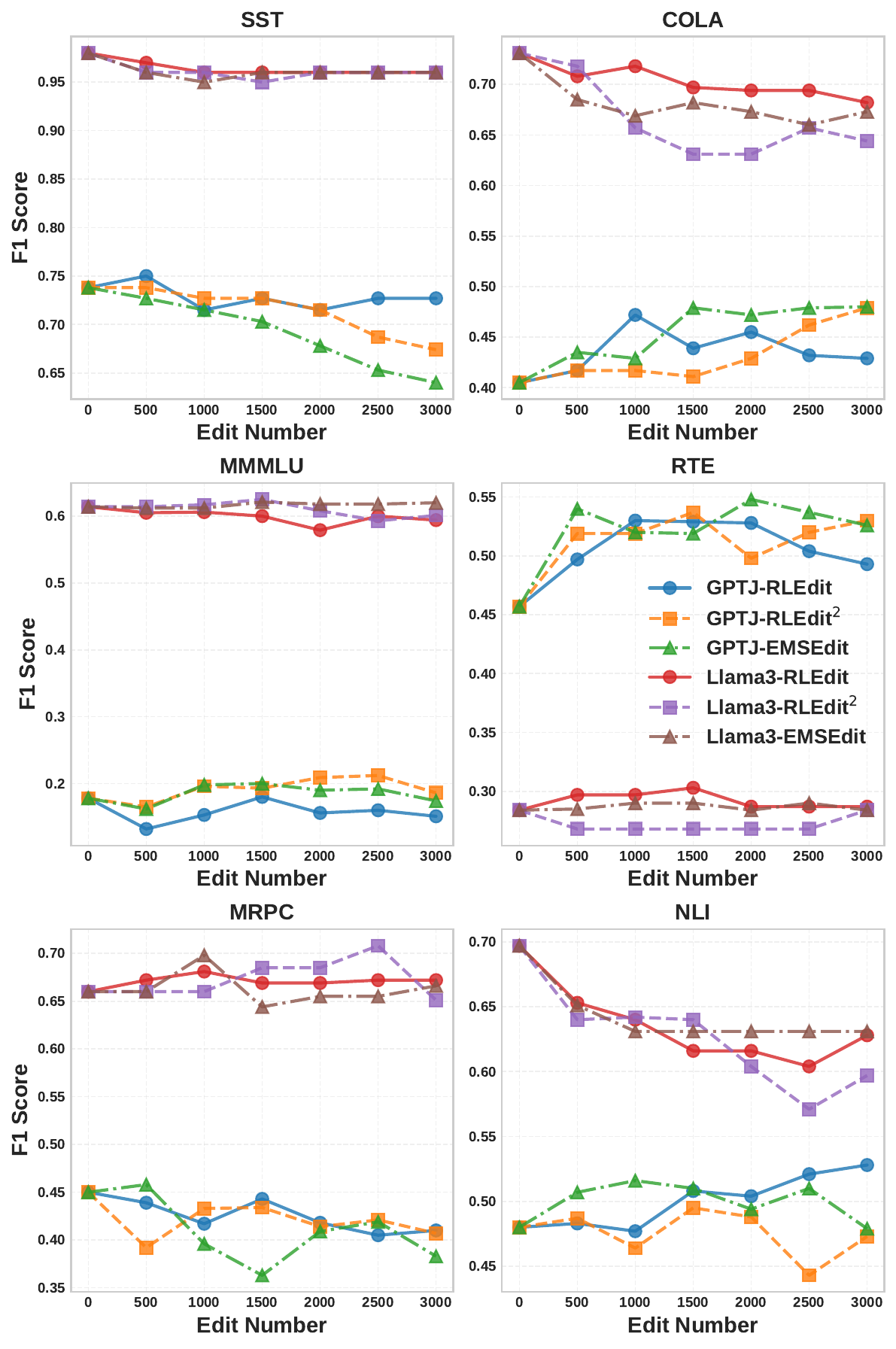}
    \caption{F1 scores of the post-edited LLMs on six tasks.}
    \label{fig:glue-res}
\end{figure}
\section{General Capability Test}\label{appendix:general}
In Section \ref{sec:main_res} and Section \ref{sec:performance_gain}, we find that although EMSEdit and MSBP exhibit strong performance in Efficacy and Generalization, they show suboptimal performance in Specificity. To further investigate the impact of EMSEdit and MSBP on the original capabilities of LLMs, we examine the performance of the edited models on general tasks. Following the work of \cite{fang2024alphaedit,li2025reinforced}, we evaluate the post-edit performance of LLMs using six natural language tasks from the General Language Understanding Evaluation (GLUE) benchmark \cite{wang2018glue}. The six tasks are: SST (The Stanford Sentiment Treebank) \cite{socher2013recursive}, MRPC (Microsoft Research Paraphrase Corpus) \cite{dolan2005automatically}, MMLU (Massive Multi-task Language Understanding) \cite{hendrycks2020measuring}, RTE (Recognizing Textual Entailment) \cite{bentivogli2009fifth}, CoLA (Corpus of Linguistic Acceptability) \cite{warstadt2019neural}, and NLI (Natural Language Inference) \cite{williams2017broad}.

We perform 30 sequential edits with a batch size of 100 on both GPT-J and LLaMA-3. The performance of the LLMs before and after editing is evaluated every 5 edits. The results are shown in Figure \ref{fig:glue-res}. It can be observed that the overall performance of the edited LLMs is comparable across the three methods: EMSEdit, RLEdit, and RLEdit$^2$. In some cases, EMSEdit outperforms the other two methods: on the CoLA and RTE tasks with GPT-J, and on the RTE, MMLU, and NLI tasks with LLaMA-3. However, on the SST task with GPT-J, EMSEdit underperforms compared to the other two methods. Despite this, these results suggest that \textbf{the introduction of MSBP do not significantly impact the ability of editing methods to preserve the original capabilities of LLMs.}
\section{Experimental Details}\label{appendix:Experimental detials}
\begin{table}[htbp]
  \centering
  \caption{Hyperparameter Settings of Model Editing Methods}
   \vspace{-0.4cm}
  \resizebox{0.5\textwidth}{!}{
    \begin{tabular}{lrrrrr}
    \toprule
    \textbf{Hyperparameter} & \multicolumn{1}{l}{\textbf{SMEdit}} & \multicolumn{1}{l}{\textbf{MALMEN}} & \multicolumn{1}{l}{\textbf{RLEdit}} & \multicolumn{1}{l}{\textbf{MALMEN$^2$}} & \multicolumn{1}{l}{\textbf{RLEdit$^2$}} \\
    \midrule
    \makecell{\textbf{Rank of linear transformation}\\ \textbf{in hyper-network}} & 1024  & 1024  & 1024  & 1024  & 1024 \\
    \midrule
    \makecell{\textbf{Number of blocks in }\\ \textbf{hyper-network}} & 4     & 4     & 4     & 4     & 4 \\
    \midrule
    \textbf{Initial learning rate} & 1e-6  & 1e-6  & 1e-6  & 1e-6  & 1e-6 \\
    \midrule
    \textbf{Meta-learning rate} & 1e-5  & 1e-5  & 1e-5  & 1e-5  & 1e-5 \\
    \midrule
    \textbf{Locality coefficient} & -     & 0.6   & 0.6   & 0.6   & 0.6 \\
    \midrule
    \textbf{Maximum meta gradient norm} & 1     & 1     & 1     & 1     & 1 \\
    \midrule
    \textbf{Regularization coefficient} & 0.5   & -     & 1e-4  & -     & 1e-4 \\
    \midrule
    \textbf{Discount factor} & -     & -     & 1     & -     & 1 \\
    \midrule
    \textbf{Backtracking depth} & -     & -     & 10    & -     & 10 \\
    \midrule
    \textbf{Memory backtracking decay factor} & -     & -     & 0.95  & -     & 0.95 \\
    \midrule
    \textbf{BP Steps} & 2     & 1     & 1     & 2     & 2 \\
    \bottomrule
    \end{tabular}}%
  \label{tab:hparams}%
\end{table}%

Experiments are conducted on A800 (80G) GPUs and a server equipped with two Intel(R) Xeon(R) Platinum 8358 CPUs @ 2.60GHz.
\subsection{Hyperparameter Setting}
To ensure a fair comparison, the hyperparameter settings of \textbf{EMSEdit}, \textbf{MALMEN}, \textbf{UltraEdit}, \textbf{MEND}, and \textbf{RLEdit} are kept largely consistent. For the variant methods \textbf{MALMEN\textsuperscript{2}}, \textbf{RLEdit\textsuperscript{2}}, and \textbf{UltrEdit\textsuperscript{2}}, all hyperparameters remain the same as their original versions, except for the BP steps. For FT, we follow the settings in EasyEdit \cite{wang2024easyediteasytouseknowledgeediting}. We present the hyperparameters of each method in Table \ref{tab:hparams}.

In our study, for \textbf{sequential editing}, all methods edit the following modules:
\begin{itemize}
    \item \textbf{GPT-J:} \texttt{mlp.fc\_in[10,12,14,16,18,20,22,24,26]}
    \item \textbf{LLaMA-3:} \texttt{mlp.gate\_proj[11--15]} and \texttt{mlp.up\_proj[18--24]}
     \item \textbf{Gemma-2:} \texttt{mlp.gate\_proj[32--40]} and \texttt{mlp.up\_proj[32--40]}
\end{itemize}

For \textbf{batch editing}, all methods edit:
\begin{itemize}
    \item \textbf{GPT-J:} \texttt{mlp.fc\_out[22--27]}
    \item \textbf{LLaMA-3:} \texttt{mlp.gate\_proj[11--15]} and \texttt{mlp.up\_proj[18--24]}
    \item \textbf{Gemma-2:} \texttt{mlp.down\_proj[32--40]}
\end{itemize}
\subsection{Details of Dataset}
\textbf{ZsRE (Zero-shot Relation Extraction) \cite{zhu2020modifying}} ZsRE is originally a relation extraction dataset. Zhu et al. \cite{zhu2020modifying} processes the training set from the KILT \cite{petroni2020kilt} version of ZsRE by evenly distributing the questions corresponding to each fact into the training and test sets, and filtering out samples that exceed the token limit. Each sample in this dataset consists of a question, a paraphrased version of the question, and a semantically inconsistent local question. The question and its paraphrase are used to evaluate the efficacy and generalization of the editing method, respectively, while the local question is used to assess specificity. The metrics are as follows:
\begin{itemize}
    \item Efficacy: It evaluates the ability of edited model $\theta_{W_1}$ to correctly predict the exact editing instance $(x,y)$:
    \begin{equation}
        \mathbb{E} [y=\arg \max\limits_{\boldsymbol{y}} \mathbb{P}_{\theta_{W_1}}(\boldsymbol{y}|x)]
    \end{equation}
    \item Generalization: It evaluates the ability of edited model $\theta_{W_1}$ to correctly predict the semantically consistent instance $(x^e, y^e)$ of the exact editing instance:
     \begin{equation}
        \mathbb{E} [y^e=\arg \max\limits_{\boldsymbol{y}} \mathbb{P}_{\theta_{W_1}}(\boldsymbol{y}|x^e)]
    \end{equation}
    \item Specificity: It evaluates the ability of edited model $\theta_{W_1}$ to correctly predict the unrelated instance $(x^u,y^u)$:
         \begin{equation}
        \mathbb{E} [y^u=\arg \max\limits_{\boldsymbol{y}} \mathbb{P}_{\theta_{W_1}}(\boldsymbol{y}|x^u)]
    \end{equation}
\end{itemize}

\textbf{CounterFact \cite{Meng2022Locating}} The CounterFact dataset originates from the PARAREL \cite{elazar2021measuring} dataset, with additional paraphrase prompts and neighborhood prompts added to evaluate generalization and specificity, respectively. Following previous work \cite{li2025reinforced}, we use efficacy, generalization, and specificity to evaluate the performance of editing methods on this dataset.
\begin{itemize}
    \item Efficacy: It evaluates whether the edited model $\theta_{W_1}$ assigns a higher probability to the edited instance $(x,y)$ compared to the old knowledge $(x,\boldsymbol{y})$:
    \begin{equation}
        \mathbb{E} [\mathbb{P}_{\theta_{W_1}}(y|x) > \mathbb{P}_{\theta_{W_1}}(\boldsymbol{y}|x)]
    \end{equation}
    \item Generalization: It evaluates whether the edited model $\theta_{W_1}$ assigns a higher probability to the semantically consistent instance $(x^e,y^e)$ of the edited instance compared to the old knowledge $(x,\boldsymbol{y})$:
    \begin{equation}
        \mathbb{E} [\mathbb{P}_{\theta_{W_1}}(y^e|x^e) > \mathbb{P}_{\theta_{W_1}}(\boldsymbol{y}|x^e)]
    \end{equation}
    \item Specificity: It evaluates whether the edited model $\theta_{W_1}$ assigns a higher probability to the unrelated instance $(x^u,y^u)$ instead of the edited instance:
 \begin{equation}
        \mathbb{E} [\mathbb{P}_{\theta_{W_1}}(y^u|x^u) > \mathbb{P}_{\theta_{W_1}}(y|x^u)]
    \end{equation}
\end{itemize}

% \section{Limitations}\label{appendix:limitations}
% Although this work demonstrates that MSBP can enhance the performance of MLME in both batch and sequential editing, our evaluation is currently limited to structure knowledge datasets. Whether this improvement extends to more complex scenarios, such as long-form knowledge, remains to be further explored. We leave this for future work.
\section{Impact Statements}\label{appendix:Impact Statements}
Model editing is an emerging field aimed at efficiently correcting errors, outdated information, and harmful content in pretrained language models. However, as a powerful technology, it can also be misused to inject malicious information. We strongly oppose such misuse and encourage the research community to develop defenses against it.


\end{document}